\documentclass[10pt,twocolumn,letterpaper]{article}

\usepackage[pagenumbers]{cvpr} %

\usepackage[dvipsnames]{xcolor}

\newcommand{\moniker}{OFER}
\newcommand{\IdGen}{\emph{Identity Generative Network}}
\newcommand{\IdRank}{\emph{Identity Ranking Network}}
\newcommand{\ExpGen}{\emph{Expression Generative Network}}

\newcommand{\flameshape}{S}
\newcommand{\arcfaceemb}{c_a}

\newcommand{\real}[1]{\ensuremath{\mathbb{R}^{#1}}}

\usepackage{enumitem}
\usepackage{multicol}
\usepackage{multirow}
\usepackage{pifont}
\newcommand{\cmark}{\ding{51}}%
\newcommand{\xmark}{\ding{55}}%
 
\usepackage[dvipsnames]{xcolor}

\def\B{{\cal B}}

\def\D{{\cal D}}

\def\J{{\cal J}}

\def\M{{\cal M}}

\def\T{{\cal T}}

\def\W{{\cal W}}

\definecolor{PColor}{rgb}{1.0,0,0} %

\definecolor{VictoriaColor}{rgb}{0,1.0,0} %

\usepackage[normalem]{ulem}       %

\definecolor{cvprblue}{rgb}{0.21,0.49,0.74}
\usepackage[pagebackref,breaklinks,colorlinks,allcolors=cvprblue]{hyperref}
\usepackage{balance}

\def\paperID{13123} %
\def\confName{CVPR}
\def\confYear{2025}

\title{\moniker: Occluded Face Expression Reconstruction}

\author{
  {Pratheba Selvaraju}\\
  \small{University of Massachusetts, Amherst}
  \and
  Victoria Fernandez Abrevaya\\
  \small{Max Planck Institute for Intelligent Systems}
  \and
   Timo Bolkart\\
  \small{Google Research}
 \and
  Rick Akkerman\\
 \small{University of Amsterdam}\\
 \and
 Tianyu Ding, Faezeh Amjadi, Ilya Zharkov\\
 \small{Microsoft Research}
}
\begin{document}
\twocolumn[{%
\renewcommand\twocolumn[1][]{#1}%
\maketitle
\begin{center}
    \centering
    \captionsetup{type=figure}
    \includegraphics[width=\linewidth]{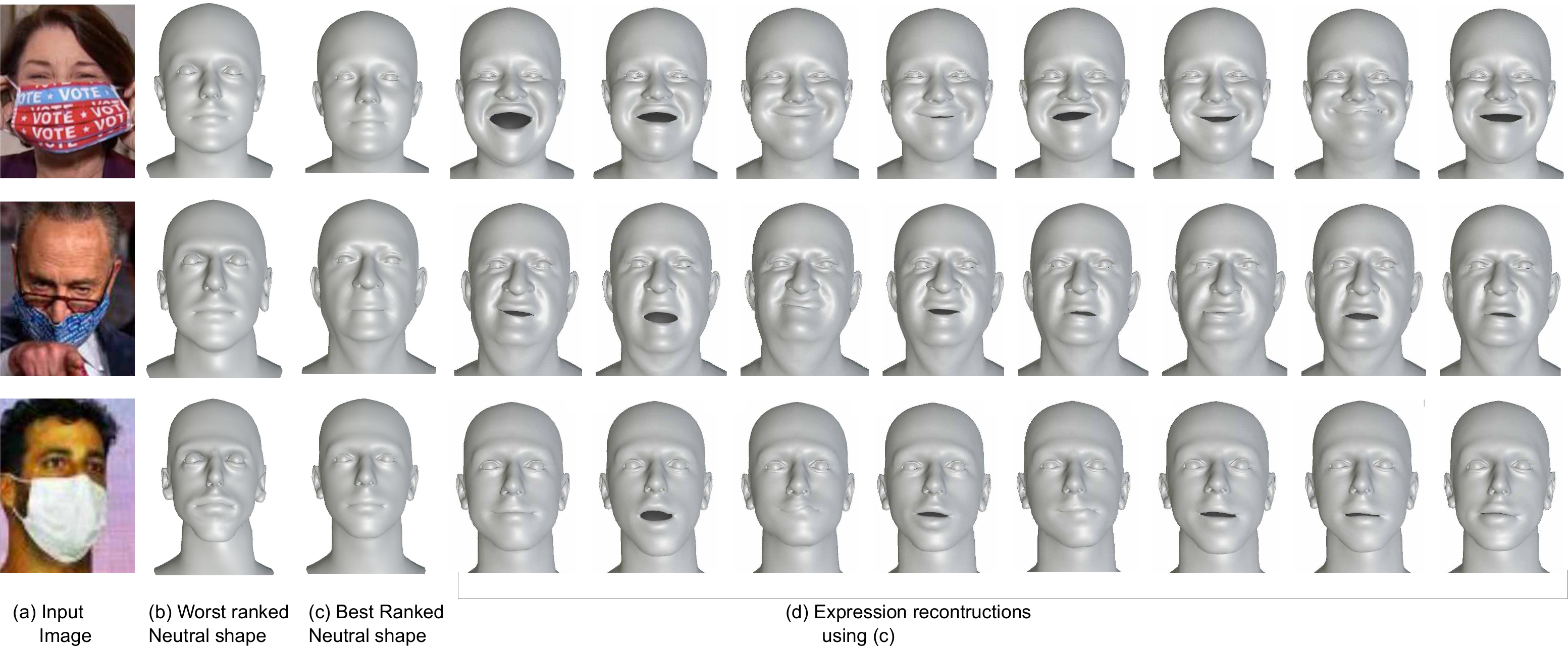}
    \caption{\label{fig:teaser}{
    \textbf{Reconstructions generated by \moniker.} Our method can reconstruct faces from a single image under hard occlusions (a), providing multiple solutions with diverse expressions that align with the input image (d). We use two diffusion models that denoise shape and expression parameters of FLAME conditioned on the image. A novel ranking mechanism selects an optimal identity (c) from the generated set of shapes, %
    on top of which the expression variants are applied to obtain the final results. 
    }}
\end{center}

}]
\maketitle

\footnotetext[1]{Project website: \url{https://ofer.is.tue.mpg.de/}}

\begin{abstract}
Reconstructing 3D face models from a single image is an inherently ill-posed problem, which becomes even more challenging in the presence of occlusions. In addition to fewer available observations, occlusions introduce an extra source of ambiguity where multiple reconstructions can be equally valid. Despite the ubiquity of the problem, very few methods address its multi-hypothesis nature. In this paper we introduce \moniker, a novel approach for single-image 3D face reconstruction that can generate plausible, diverse, and expressive 3D faces, even under strong occlusions. Specifically, we train two diffusion models to generate the shape and expression coefficients of a face parametric model, conditioned on the input image. This approach captures the multi-modal nature of the problem, generating a distribution of solutions as output. 
However, to maintain consistency across diverse expressions, the challenge is to select the best matching shape. To achieve this, we propose a novel ranking mechanism that sorts the outputs of the shape diffusion network based on predicted shape accuracy scores. We evaluate our method using standard benchmarks and introduce CO-545, a new protocol and dataset designed to assess the accuracy of expressive faces under occlusion. 
Our results show improved performance over occlusion-based methods, while also enabling the generation of diverse expressions for a given image.

\end{abstract}

\section{Introduction}
\label{sec:intro}

3D face reconstruction from a single image 
is crucial for creating life-like digital avatars and is used in numerous applications such as illumination-invariant recognition~\cite{Zou2007IlluminationInvariant}, medical imaging~\cite{Pan2017Medical3D}, and telepresence~\cite{Lee2023Farfetch}. 
The task is an inherently challenging inverse problem, with difficulties posed by
depth ambiguity, %
variations in lighting, diverse facial expressions, and pose~\cite{diao2024survey}. 

The problem becomes even
more challenging when images are subjected to \emph{occlusions}, as shown in \cref{fig:teaser}. 
The main difficulty arises from the face being only partially visible, 
introducing an additional %
source of ambiguity:
the occluded areas can now correspond to an infinite number of valid shapes, making it a \emph{multi-hypothesis reconstruction} problem.  
Occlusions are very common in images captured in uncontrolled environments %
due to factors such as hair, accessories, medical masks, or even strong profile poses and head rotations (\eg \cref{fig:aflw2000}).
Despite its prevalence, 
reconstruction under such conditions %
has seldom been addressed. 

\begin{figure}
    \centering
    \includegraphics[width=1.0\linewidth]{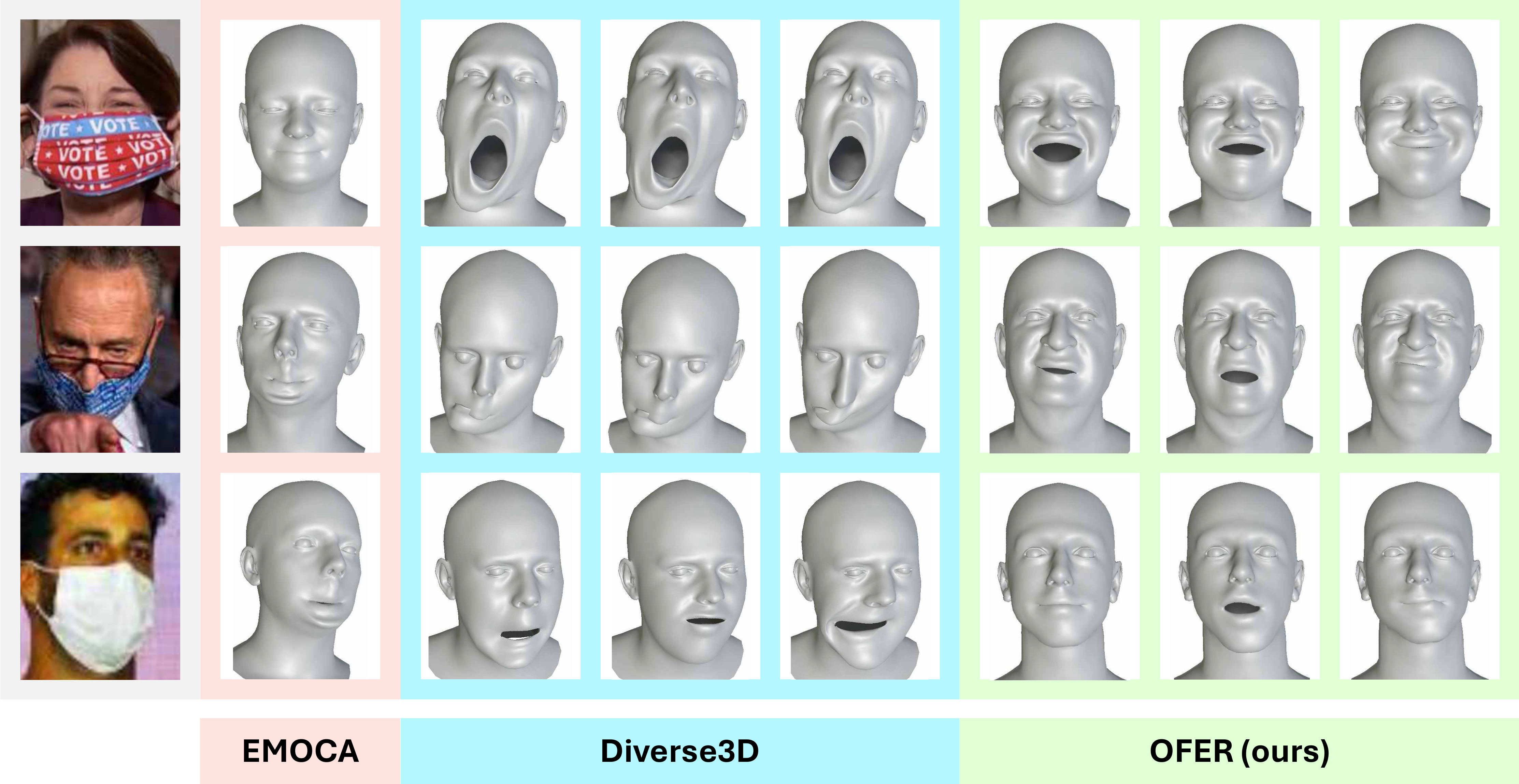}
    \caption{\label{fig:exp_comparison}{
    \textbf{
    Comparison against %
    expression reconstruction 
    methods. } 
    We show results from EMOCA~\cite{danecek2022emoca} (pink); samples generated by Diverse3D~\cite{dey2022generating} (blue); and samples generated by our method (green). 
    EMOCA can only reconstruct a single solution
    due to its deterministic nature, while Diverse3D shows non-plausible faces. OFER (our method) generates diverse 3D faces with plausible expressions.}}
\end{figure}

3D face reconstruction typically involves recovering the parameters of a statistical 3D facial model, such as the 3D Morphable Model (3DMM)~\cite{Blanz19993dmm}, either via model fitting~\cite{Blanz19993dmm, aldrian2013InverseRO, gecer2019ganfit} %
or regression~\cite{danecek2022emoca, deng2019accurate, feng21deca, tran2018extreme}. While these works may %
handle milder occlusions depending on their training data, their performance deteriorates significantly under more severe obstructions %
(see \eg~\cref{fig:exp_comparison}). A few methods have been specifically proposed to address the challenge of occluded faces~\cite{egger2016occlusion, egger2018occlusion, li2023robust, dey2022generating}. Most %
offer a unique solution due to their deterministic nature, ignoring the multi-hypothesis aspect %
of the problem~\cite{egger2016occlusion, egger2018occlusion, li2023robust}. %
An exception to this is Diverse3D~\cite{dey2022generating}, which employs %
Determinantal Point Process (DPP)~\cite{Kulesza2012DeterminantalPP} to sample a diverse set of outputs. While DDP is designed to %
capture the diversity of the data, %
it fails to adequately represent %
the underlying distribution of facial geometry,  %
often resulting in unrealistic and exaggerated reconstructions as shown in \cref{fig:exp_comparison}.

To ensure plausible face reconstructions while addressing the challenges of diverse sampling, we introduce \moniker, a novel method for reconstructing 3D faces under occlusions. 
 At its core, \moniker{} employs two denoising diffusion probabilistic models (DDPMs)~\cite{ho2020denoising} %
 to generate the shape and expression coefficients of the FLAME~\cite{li2017flame} parametric face model, conditioned on the input image. %
 The ability of diffusion models to learn the underlying data distribution enables \moniker{} to produce multiple plausible hypotheses.
 We leverage two pre-trained face image recognition networks as image encoders, %
 and following MICA~\cite{zielonka2022mica}, train our models using only a small dataset of paired 2D-3D data.

Generating diverse solutions is crucial for reconstructing faces under occlusions. 
Equally important is selecting a unique, consistent shape that best represents the face in the image across all generated expressions. This is achievable since identity is generally better defined under occlusions. 

To achieve this, we propose %
a novel \textbf{ranking mechanism} that evaluates and ranks the samples generated by the shape diffusion network, effectively filtering out poor quality samples. 
Specifically, given an input image along with N generated shape samples, %
the ranking network scores and ranks the candidates, %
selecting 
the top-ranked one. 
To the best of our knowledge, such a ranking mechanism as selection process for diffusion models has not been considered before. %

To evaluate our approach, we introduce a new dataset and evaluation protocol, CO-545, derived from the CoMA dataset~\cite{ranjan2018coma}. %
CO-545 includes 545 ground-truth pairs of occluded images and corresponding FLAME non-occluded 3D vertices. Experimental results show that our method effectively generates multiple hypotheses for single in-the-wild occluded images, achieving superior quality and diversity %
on CO-545. Additionally, when evaluated on the NoW benchmark~\cite{Sanyal2019RingNet}, our method demonstrates improved performance with the ranking-selection mechanism.

In summary, our contributions are the following:
\begin{itemize}
    \item A new method for occluded 3D face reconstruction using DDPMs, %
    which outputs multiple 3D face shape and expression hypotheses, %
    overcoming the limitations of deterministic methods. %
    
    \item A novel ranking mechanism that scores and selects the optimal 3D face shape %
    from reconstruction candidates generated by the shape network.%

    \item A new validation dataset, CO-545, for the quantitative evaluation of occluded face reconstruction,  addressing the lack of existing evaluation protocols.
\end{itemize}
\section{Related Work}

\paragraph{3D face reconstruction from a single image.}
3D face reconstruction from a single image has been a key research area for several decades~\cite{egger2020survey}, with approaches broadly categorized into model-free~\cite{deng2020retinaface,feng2018prn,Ruan2021SADRNetSD,dou2017endtoend,alp2017densereg,jung2021ffd,sela2017unrestricted, zeng2019DF2Net} and model-based methods~\cite{zielonka2022mica, feng21deca, danecek2022emoca, zhang2023tokenface, retsinas2024smirk}. Model-free approaches estimate 3D geometry directly from images, while model-based ones recover the low-dimensional parameters of a statistical model of the 3D face such as BFM~\cite{paysan2009bfm} or FLAME~\cite{li2017flame}. To overcome the lack of large-scale datasets with paired images and 3D models, recent trends have shifted towards self-supervised learning~\cite{feng21deca, danecek2022emoca, Tewari2017MoFA, deng2019accurate, retsinas2024smirk}, using landmark re-projection and/or photometric error. %
These works can yield suboptimal results when landmarks are missing or when color information is compromised due to occlusions. Alternatively, MICA~\cite{zielonka2022mica} uses a small dataset of paired 2D-3D data to map an image embedding from a face recognition network~\cite{Deng2018ArcFaceAA} to the FLAME shape parameters. This technique achieves state-of-the-art results for neutral face shape reconstruction but does not support expressive faces. 

\paragraph{3D face reconstruction from occluded images.}
A few works have specifically addressed the problem of reconstructing faces under occlusion. Egger et al.~\cite{egger2016occlusion, egger2018occlusion} proposed a probabilistic optimization approach that simultaneously solves for model parameters and segmentation regions using an expectation–maximization approach. FOCUS~\cite{li2023robust} follows a similar idea within a self-supervised learning strategy. These methods prioritize obtaining valid reconstructions for the non-occluded regions 
but do not address the ambiguity of the problem, which requires a distribution of solutions.
Recently, Diverse3D~\cite{dey2022generating} tackled this by employing a mesh-based variational autoencoder (VAE) for shape completion, combined with Determinantal Point Process (DPP) ~\cite{Kulesza2012DeterminantalPP} for sampling diverse solutions in expression space. %
However, the method often results in unrealistic and exaggerated reconstructions (see~\cref{fig:exp_comparison}), since DDP is not designed to capture the statistical properties of %
the data.

\paragraph{\textbf{Learning to Rank.}}
Ranking plays an essential role in Information Retrieval~\cite{Liu2009LearningtoRank}, %
commonly used for sorting documents by relevance~\cite{ko2022survey,han2020learning}, image search~\cite{Parikh2011RelAttr}, and recommendation systems~\cite{Nguyen2014LearningTR}. 
Learning to Rank methods~\cite{Liu2009LearningtoRank} perform this task by making use of supervised machine learning.
RankGAN~\cite{Lin2017AdversarialRF} uses ranking as a rewarding mechanism to train a language generator to produce higher-quality descriptions.
Some recent works such as T5~\cite{zhuang2023rankT5} focus on text ranking with large language models, and incorporate ranking losses as a fine-tuning tool to optimize model performance.  
To the best of our knowledge, ranking has not been explored in the context of diffusion models to select an optimal sample.

\section{Method}
\label{sec:method}

Our method takes as input a single-view image of an occluded face and generates a \emph{set} of 3D faces as output. The goal is to produce reconstructions that explore a \emph{diverse range of expressions} in the occluded areas, while accurately capturing %
the visible regions of the input image. 

\begin{figure}
    \centering
    \includegraphics[width=1\linewidth]{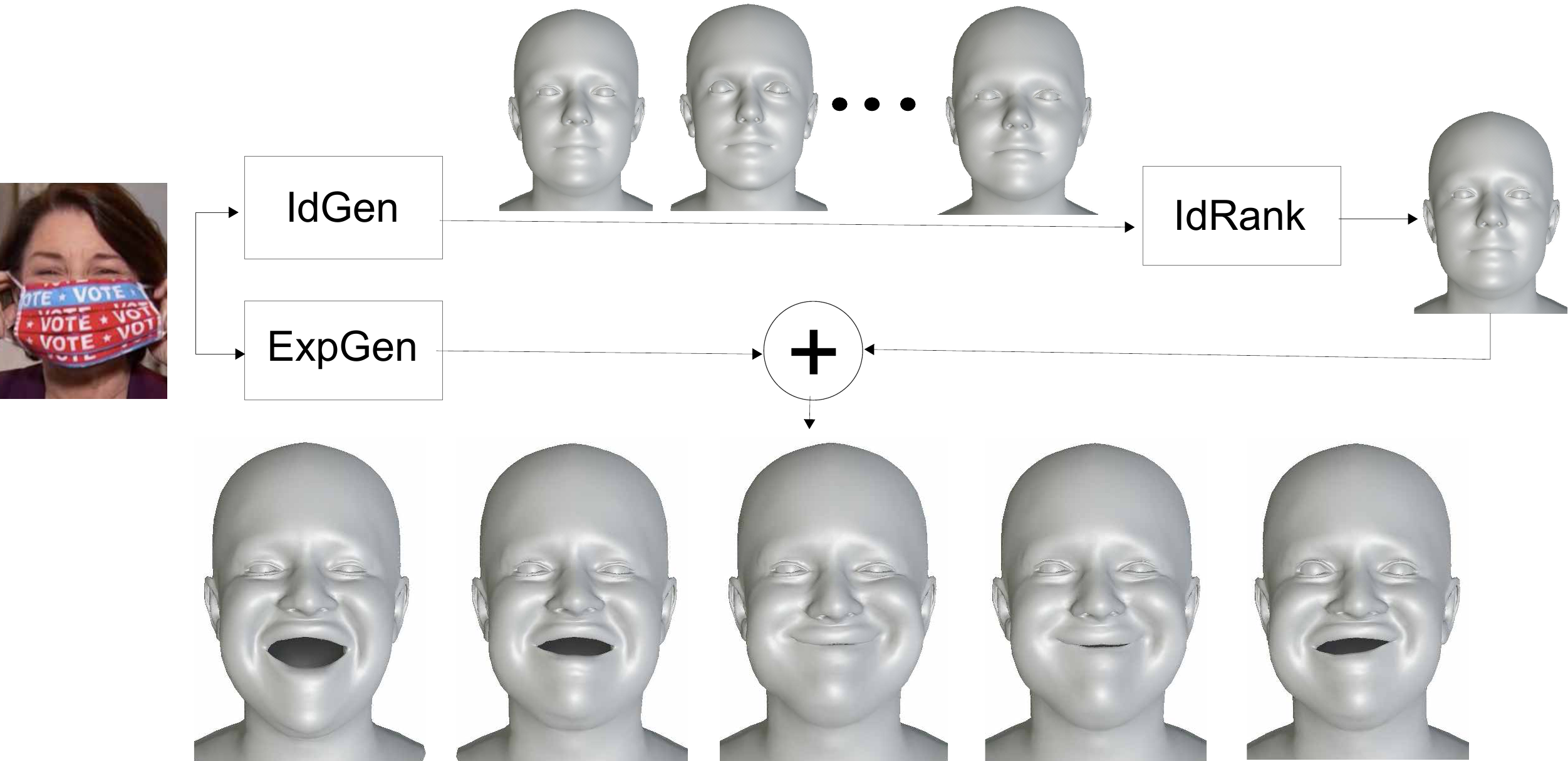}
    \caption{\textbf{Overview of \moniker.} Given an input image, the \IdGen{} (IdGen) samples N shape parameters. The reconstructed shapes are then passed to the \IdRank{} (IdRank) to select a unique identity. Finally, the \ExpGen{} (ExpGen) generates N expression parameters, which are combined with the selected shape 
    to output \emph{diverse and expressive face reconstructions} (bottom row).}
    \label{fig:overall-arch}
\end{figure}

\moniker{} is structured around three key components. 
First, the \IdGen{} (IdGen, \cref{fig:Architecture_Generative_Network}
and \cref{ssec:idgen}), 
a DDPM conditional diffusion model, outputs a set of FLAME ~\cite{li2017flame} shape coefficients, %
recovering a distribution of plausible neutral 
3D faces. 
Next, the \IdRank{} (IdRank, \cref{fig:RankArchitecture} and \cref{ssec:idrank}), a small MLP, evaluates and ranks the shape samples given by the previous network. %
This step selects %
a unique shape coefficient that best explains the image. 
Finally, the \ExpGen{} (ExpGen, \cref{fig:Architecture_Generative_Network}
and \cref{ssec:exp_network}), another conditional DDPM %
network, generates a diverse set of FLAME expression coefficients. %
We use default DDPM framework~\cite{ho2020denoising} with DDPM sampler with 1000 sampling time steps for both IdGen and ExpGen.
The three networks are conditioned on the same input image. 
By combining the selected shape coefficient %
with the set of expression hypotheses, %
we obtain the final set of 3D reconstructions.
The overall architecture of \moniker{} is shown in ~\cref{fig:overall-arch}.  We detail each of its components in the following sections.

\subsection{Preliminaries}
\paragraph{FLAME Model.}
FLAME \cite{li2017flame} is a parametric 3D face model combining separate shape and expression spaces to form a 3D head mesh $M$ with $n=5023$ vertices. Given the shape $S \in \real{|S|}$, expression  $E \in \real{|E|}$, and pose $P \in \real{|P|}$ parameters, FLAME produces a mesh $M$ as 
\begin{equation}
\label{eq:flame}
M(S, E, P) = \text{LBS} \big( \T(S, E, P), \J(S), P, \W \big), 
\end{equation}
where $\text{LBS}$ is the linear blend skinning function using weights $\W \in \real{4 \times n}$ and joint regressor $\J(S) \in \real{3K}$, and 
$T(S, E, P) \!= \T + \B_S(S) + \B_E(E) + \B_P(P)$
deforms a template mesh $\T$ using the shape $\B_S$, expression $\B_E$ and pose-corrective $\B_P$ blendshapes. FLAME uses $|S|=300$ dimensions for shape and $|E|=100$ dimensions for expression. 

\paragraph{Denoising Diffusion Probabilistic Model (DDPM).}
The underlying mechanism of DDPM %
is the %
transformation of an unknown data distribution, $p(X) \in \mathcal{R}^d$ into a simple known distribution. Here, $X$ represents the data from the underlying distribution. 
This transformation is achieved by iteratively applying a transition kernel $q$ via a Markov chain process with infinitesimal time steps, ensuring a stationary distribution at each time step $t$ i.e $x_t \sim q(x_t|x_{t-1}), \forall t > 0$ where $x \in X$. DDPM models these transformations by parameterizing %
a neural network %
to captures complex dependencies in the data by modeling the sequential evolution of the data distribution. In the work of Sohl et al.\cite{sohl2015deep}, the known distribution is set as gaussian with a decaying variance schedule $\beta_t \in \mathcal{R}$ such that $ q(x_t|x_{t-1} = \mathcal{N}(x_t; \sqrt{1-\beta_t}x_{t-1}, \beta_tI); q(x_T) = \mathcal{N}(x_T; 0, I)$ due to its simplicity and tractability. Denoising in DDPM involves reversing this process by starting with unit gaussian noise and attempting to model the underlying data distribution $p'(X)$.

\paragraph{Learning-based Ranking.}
 Ranking is an important technique used in the information retrieval domain to retrieve relevant documents given a query. %
There are three primary learning-based ranking techniques~\cite{Liu2009LearningtoRank}: pointwise, pairwise and listwise. The pointwise model takes in a single query($q)$-document($d_i$) pair and gives the relevance score ($s_i$) of each in isolation, which requires the ground-truth score: $\Bar{r}(q,d_i) = s_i$. Pairwise ranking methods compare two documents against each other based on importance and relevance to the query: $\Bar{r}(q, d_i, d_j) = P(d_i \triangleright d_j)$. Finally, listwise ranking~\cite{rahimi2019listwise} generates an optimal order of a list of documents by calculating the importance score of each: $\Bar{r}(q,d_i, \ldots, d_n) = (r_1,\ldots, r_n)$, which is the method we adopt in our network.

\subsection{Identity Generative Network (IdGen)}
\label{ssec:idgen}
\begin{figure}[t!]
    \centering
    \includegraphics[width=1.0\linewidth]{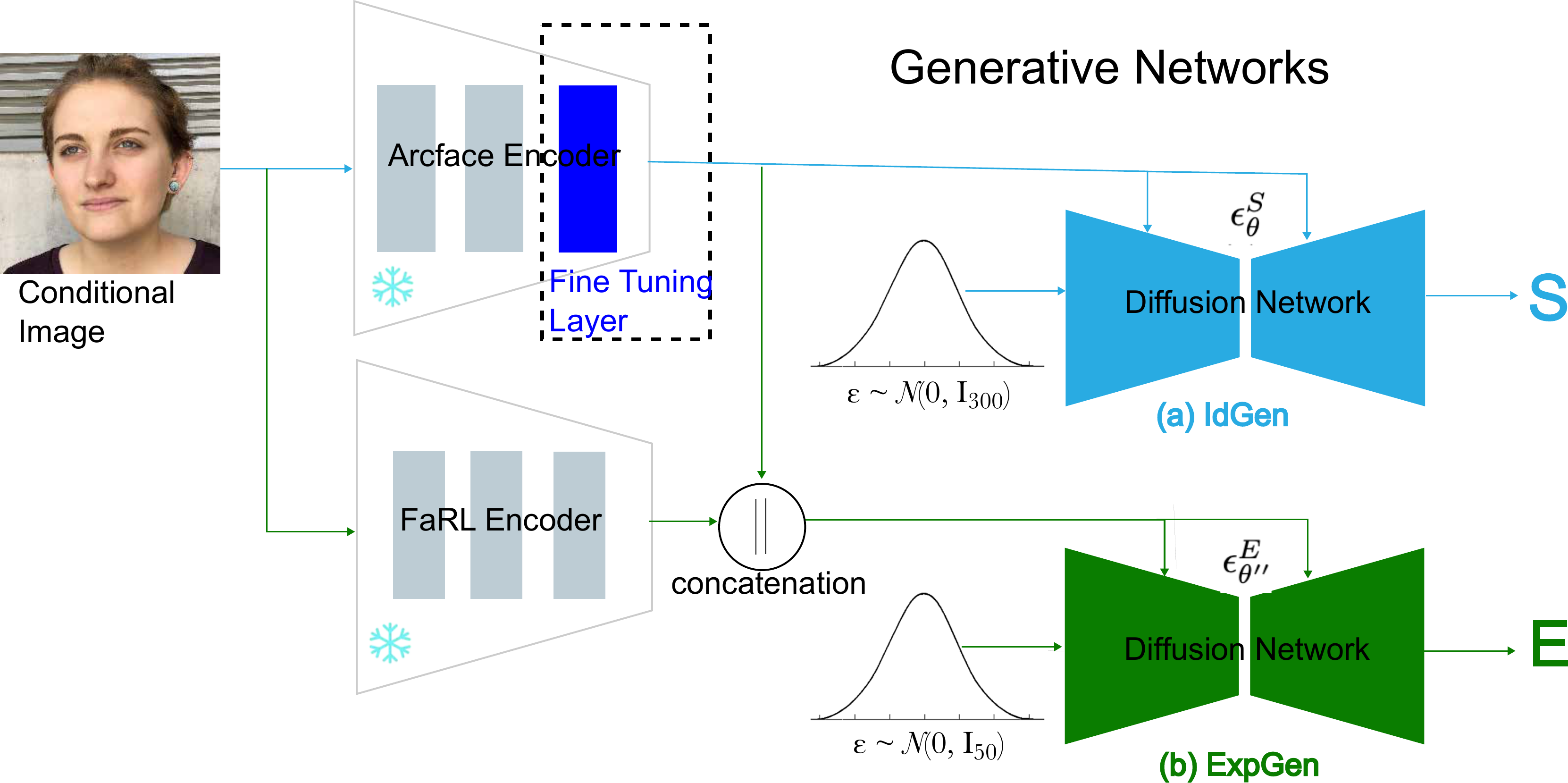}
    \caption{\textbf{Overview of the identity and expression %
    generative networks} %
    (IdGen, in {\color{cyan}blue}, and ExpGen, in {\color{ForestGreen}green}).
    For {\color{cyan}IdGen}, the input image is encoded into a 512-dimensional embedding using ArcFace~\cite{Deng2018ArcFaceAA}. %
    This serves as a condition for the 1D U-Net diffusion network, which is trained to denoise 300-dimensional noise into FLAME shape coefficients, S.
    For {\color{ForestGreen}ExpGen}, the input image is encoded into a 1024-dimensional embedding using the FaRL~\cite{zheng2022general} and ArcFace~\cite{Deng2018ArcFaceAA} encoders. The embedding 
    serves as a condition for the 1D U-Net diffusion network, which is trained to denoise 50-dimensional noise into FLAME expression coefficients, E.
    }
    \label{fig:Architecture_Generative_Network}
\end{figure}

The first step of \moniker{} is to generate a set of FLAME shape coefficients, $S$. 
The identity %
network $\epsilon^S_{\theta}$ (\cref{fig:Architecture_Generative_Network}, top)
consists of 
a 1-dimensional U-Net~\cite{Ronneberger2015Unet} %
with self-attention layers 
that is trained to transform unit Gaussian noise $s_T \sim \mathcal{N}(0, I_{300})$ into $s_0 = S$ 
through reverse time-steps $t = {T..0}$.

The input image is used as condition %
by encoding it into a 512-dimensional feature vector $\arcfaceemb \in \real{512}$
using the ArcFace \cite{Deng2018ArcFaceAA} encoder. %
 Following  MICA \cite{zielonka2022mica}, we freeze all layers of the ArcFace encoder except the last,
 such that the encoded features can be fine-tuned during training.  
During sampling, we generate a set of $N$ noise vectors $\{s_T^i \in \mathbb{R}^{300}\}_{i=1}^N$ and iteratively denoise them using $\epsilon^S_{\theta}$ to obtain the FLAME shape coefficients, $\{s_0^i=\hat{\flameshape}_i \in \mathbb{R}^{300}\}_{i=1}^N$. %
We optimize the parameters $\theta$ by minimizing the following loss function $L_{\varepsilon}$: 
\begin{align}
\label{eq:idgen_loss}
     &L_{\varepsilon}(\theta) %
     = \mathbb{E}_{t} \Bigl[ \mathbb{E}_{s_0,\varepsilon}[| \epsilon^S_{\theta}(s_t)  - \varepsilon_t|] \Bigr]. 
\end{align}
where $\varepsilon_t$ is fixed Gaussian noise with variance proportional to the linear time step $t$~\cite{ho2020denoising}. 
Due to the %
nature of diffusion models, each generated sample %
$\hat {S_i}$ %
exhibits plausible variations in the occluded regions, %
capturing the variance in the input data. %

\subsection{Identity Ranking Network (IdRank)}
\label{ssec:idrank}
Our goal is to generate a set of 3D faces with diverse expressions that align with the input occluded image. 
 While expressions can vary, we would like here to preserve a unique and consistent identity across all samples.
This approach is motivated by three main observations. 
First, we noticed that shape coefficients display much less variability than expression coefficients, as images typically contain more cues for the expression-independent facial features. 
Second, we noticed in our experiments that, %
when computing the reconstruction error for the N generated samples, the minimum value is on-par %
to state-of-the-art methods
(see supplementary material). 
Since identity is typically more defined than expression, a method capable of selecting this optimal sample would be advantageous. 
Finally, even in cases of severe occlusions where identity features may be less discernible, having a method that effectively filters out poorly generated samples is important for practical applications. 

Towards this end, we introduce IdRank, %
a network specifically designed 
to rank the outputs of IdGen in order to select those that best preserve the facial structure of the input image.

\begin{figure}[htb!]
    \centering
    \includegraphics[width=1.0\linewidth]{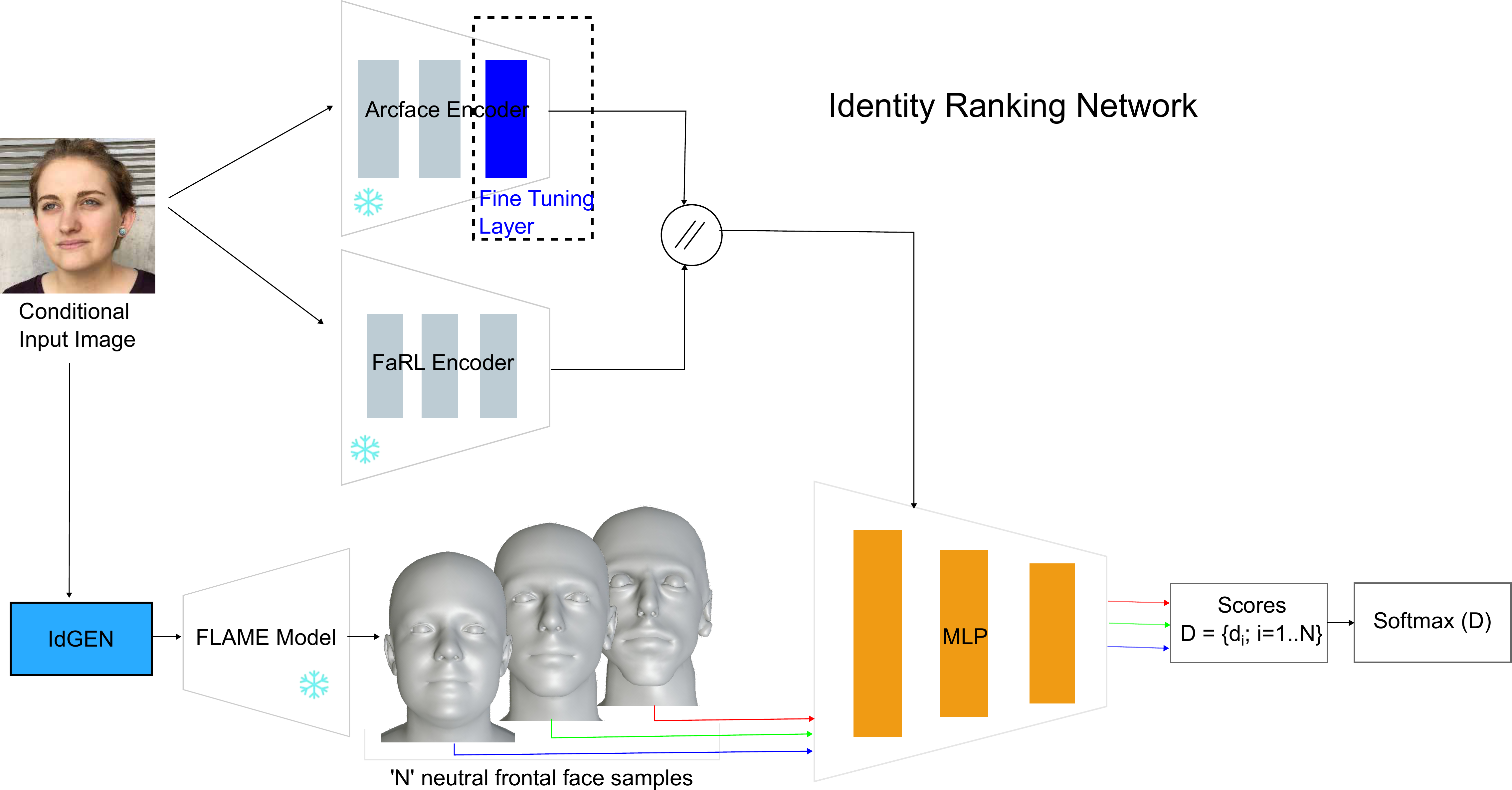}
    \caption{\textbf{Identity Ranking Network}. 
    Given the N shape coefficients from IdGen, we reconstruct the neutral meshes using FLAME. Each mesh is passed through a %
    5-layer MLP to compute a score, %
    conditioned on the input image using the ArcFace~\cite{Deng2018ArcFaceAA} and FaRL~\cite{zheng2022general} encoders. The N scores are then converted into probabilities using softmax. The ranking order of the sorted scores %
    is compared against the ranking of the sorted reconstructed errors, %
    and the network is trained to match them. %
    }
    \label{fig:RankArchitecture}
\end{figure}

\paragraph{Ranking framework.} %
{\label{para:problem_setting}}
 The overall architecture of IdRank is shown in~\cref{fig:RankArchitecture}. Given a pool of N candidate shapes generated by IdGen, the goal is to select the one closest in distance to the ground-truth neutral shape (available during training). To achieve this, we train a \emph{ranking} network %
$R_{\theta'}$ that goes from a list of $v$ face vertices (derived from the shape coefficients) to a scalar $p$ representing the probability that the sample matches the ground truth. We then apply the network $R_{\theta'}$ over the $N$ samples, ranking each based on their scores and selecting the highest-scored sample.

Before passing the mesh to the network, we remove the vertices corresponding to the back of the head since they are rarely visible in the image, and their inclusion in the error computation might lead to spurious results.

The number of candidates N is an important hyper-parameter, as it defines a trade-off between computational complexity and reconstruction error --a large value of $N$ means that the actual ground-truth shape has more chances of being generated, but incurs in a significantly higher computational cost. We empirically choose here $N=100$ (see supplementary material).

\paragraph{Generating ground-truth data.} 
During training, we generate the ground-truth data for ranking in an online fashion as follows. Given an image $I$ together with its ground-truth FLAME mesh $M_{GT} \in \real{5023}$, we first sample $N$ shape parameters $\{ \beta_i \}_{i=1}^N$ using IdGen (with gradient back-propagation disabled). We then reconstruct the neutral face vertices using \cref{eq:flame}, %
yielding a set of mesh candidates $\M = \{ M_i=M(\beta, \mathbf{0}, \mathbf{0}) \in \real{5023} \}_{i=1}^N$. From the set $\M$ as well as the ground-truth mesh $M_{GT}$ we retain only the frontal vertices using a pre-computed mask, resulting in $\M^{\text{frontal}} = \{ \hat{M}_i \in \real{n'} \}_{i=1}^N$ and $M^{\text{frontal}}_{GT} \in \real{n'}$ ($n' < 5023$). Finally, we compute the error between each $\hat{M}_i$ and $M^{\text{frontal}}_{GT}$ to produce the following set:
\begin{equation}
    \D_{GT} = \{ | \hat{M}_i - M^{\text{frontal}}_{GT} | \}_{i=1}^N.
\end{equation}
with $ | \cdot |$ the L1 norm. 

We empirically observed that removing the mean $\mu_{\mathcal{M}} = \frac{1}{N} \sum_{i=1}^N \hat M_i$ from the set $\M^{\text{frontal}}$ reduces redundancy and helps convergence of the network. Hence, we transform $\M_{\text{frontal}}$ into a zero-centered set of residual meshes $\M_{\text{centered}}$ via: %
\begin{equation}
    \M_{\text{center}} = \{ M'_i = (\hat{M}_i - \mu_{\mathcal{M}}) \}_{i=1}^N.
\end{equation}

The final training set for image $I$ is defined as the list of pairs ${\label{eq:mean_residual}}\mathbf{X} = [ (\mu_{\mathcal{M}}, M'_i) ]_{i=1}^N$, obtained by sorting the values in $\D_{GT}$ in ascending order. %

\paragraph{Learning to rank facial shapes.} 
The ranking network %
$R_{\theta'}$ is a small MLP %
that takes as input a single pair $(\mu_{\mathcal{M}}, M_i')$ and is trained to predict a sample score $d_i$. %
It is conditioned on a 1024-dimensional feature vector $c_R = (c_a \| c_f)$, formed by concatenating features from the ArcFace ($c_a$)~\cite{Deng2018ArcFaceAA} and FaRL ($c_f$) encoders~\cite{zheng2022general}, designed for facial analysis tasks. %
The network is run $N$ times producing $\D = \{d_1, \dots, d_N \}$ scores, which are then passed through a softmax operator to transform them %
into probabilities, $P = \mathrm{softmax}(\D) =  \{ p_i = \frac{e^{d_i}}{\sum_{j=1}^N e^{d_j}}\}_{i=1}^N$. The ground-truth distance values $\D_{GT}$ are also passed through a softmax operator, $\D_{GT}' = %
\mathrm{softmax}(\D_{GT})$. Finally we compute the cross-entropy loss between the two distributions $P$ and $\D_{GT}'$:
\begin{equation}
    L_{\D_{GT}', P}(\theta') = - \sum_{i=1}^N {\D_{GT}'}(i) \log (p_i). 
\end{equation}

\paragraph{Inference.}
During inference, we select the sample with the highest probability score 
as the optimal sample, denoted as $S_R$. While multiple samples may have equal probability scores, their ordering in such cases would be rather random. In that case, we choose the first-ranked sample from this list of equally high probability samples.

\subsection{Expression Generative Network (ExpGen)}
\label{ssec:exp_network}

The expression network $\epsilon_{\theta''}^E$  architecture (\cref{fig:Architecture_Generative_Network}, bottom)
closely follows that of IdGen, %
employing the conditional DDPM framework to generate the first 50 components of the FLAME expression coefficients. 
It is conditioned on a 1024-dimensional feature vector, similar to IdRank. %

During sampling, the network initializes from unit Gaussian noise $\varepsilon_T \in \mathbb{R}^{50}$ %
and iteratively denoises towards the FLAME expression parameters, %
$e_0 \in \mathbb{R}^{50}$.
The parameters $\theta''$ are optimized by minimizing the loss function \cref{eq:expgen_loss}
\begin{align}
\label{eq:expgen_loss}
     &L_{\varepsilon}(\theta'') %
     = \mathbb{E}_{t} \Bigl[ \mathbb{E}_{e_0,\varepsilon}[| \epsilon^E_{\theta}(e_t)  - \varepsilon_t|] \Bigr]. 
\end{align}

\section{Experiments and Evaluations}
\label{sec:results}

In this section, we present %
qualitative and quantitative evaluation 
results for \moniker. %
We begin by introducing the datasets used for training and testing, including the proposed CO-545 dataset, which we specifically designed to evaluate performance on occluded images (\cref{ssec:datasets}). Next, we provide %
evaluation metrics and experimental results on IdGen and IdRank (\cref{ssec:results_idgen}), followed by a similar analysis for the final reconstructions with ExpGen (\cref{ssec:results_exp}). %

\subsection{Datasets}
\label{ssec:datasets}

\paragraph{Training data.} We train IdGen with the Stirling~\cite{feng2018evaluation}, FaceWarehouse~\cite{cao2013facewarehouse}, LYHM~\cite{hang2020lyhm}, and Florence~\cite{bagdanov2011florence} datasets, which contain 2D-3D ground-truth pairs of images and corresponding registered FLAME shape %
coefficients. The same four datasets are used to train  {IdRank}. For {ExpGen} we use the FaMoS (Facial Motion across Subjects)~\cite{bolkart2023Famos} dataset, which offers paired image and FLAME 3D meshes of faces in motion. The shape and expression coefficients were provided by the FaMos authors. %

\paragraph{The CO-545 dataset.} To evaluate reconstruction of expressive faces under occlusions, we propose a new dataset built on top of the CoMA dataset~\cite{ranjan2018coma}. The CO-545 validation dataset comprises 545 frontal face images (no profile or rotated head views) synthetically occluded (hand, face, and random) across 11 subjects taken from COMA~\cite{ranjan2018coma}. %
For each image we identify %
the sets of occluded and unoccluded vertices, and evaluate them using %
separate metrics. %
More details can be found in the supplementary material.

\paragraph{Testing data.} We evaluate the final expressive reconstructions using the dataset provided by Dey et al.~\cite{dey2022generating} (``Dey Dataset''), as well as the CO-545 dataset. Additionally, we test on the real-world hard occlusion images dataset provided by Erakiotan et al.~\cite{erakiotan2021recognizing}, which does not have any 3D ground truth data. We additionally evaluate the quality of neutral shape reconstructions using the occluded and unoccluded subsets of the NoW benchmark~\cite{Sanyal2019RingNet}. Note that we do not have access to the occluded/unoccluded split for the testing set, and hence provide results mainly for the validation set. Additional comparisons on the test set are included in the supplementary material. %

\subsection{Evaluation on IdGen and IdRank}
\label{ssec:results_idgen}

\begin{figure}
    \centering
    \includegraphics[width=1.0\linewidth]{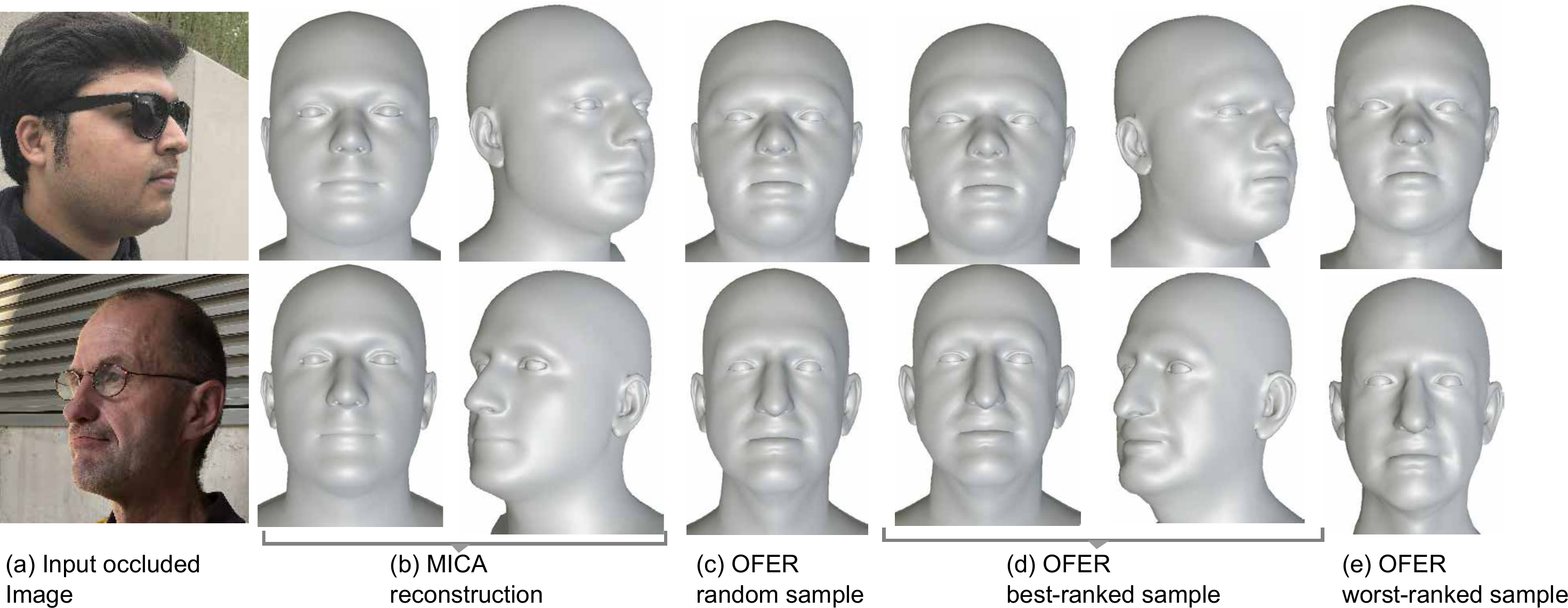}
    \caption{\textbf{Neutral face reconstruction on occluded images}. For (a) the given occluded input image, (b) shows the reconstructed shape provided by MICA~\cite{zielonka2022mica}; (c) is one of the generated samples from our method; (d) and (e) are the best and worst-ranked samples, respectively, as selected by the ranking network.}
    \label{fig:qual_Neutral_face_ranking}
\end{figure}

\begin{figure}
    \centering
    \includegraphics[width=0.7\linewidth]{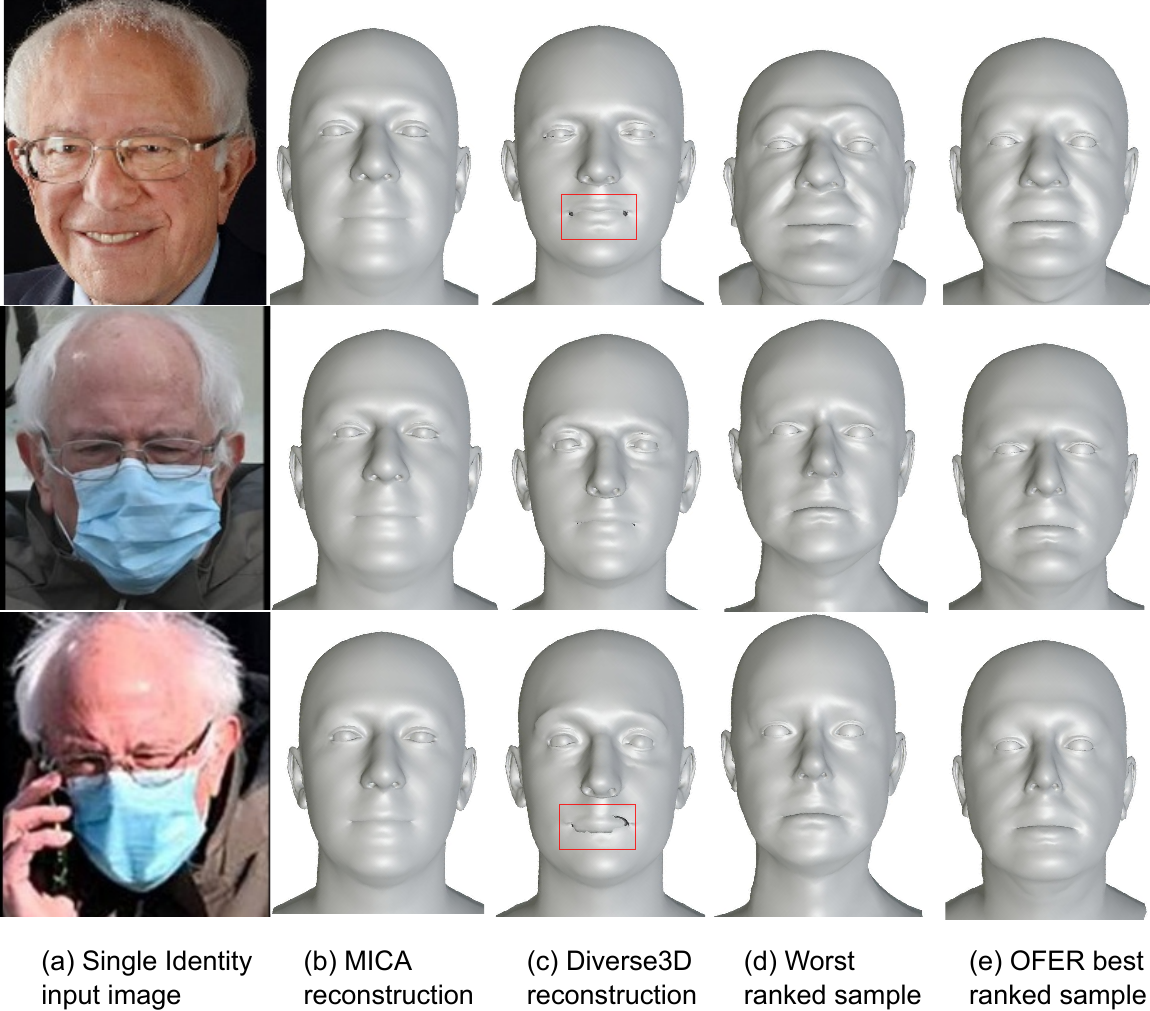}
    \caption{\textbf{Neutral face reconstruction on a single identity}. For (a) the given input image, (b) shows the reconstructed shape provided by MICA~\cite{zielonka2022mica}; (c) shows the reconstructed shape provided by Diverse3D~\cite{dey2022generating}; (d) and (e) are the worst and best-ranked 
    samples as selected by the ranking method. %
    }
    \label{fig:Neutral_face_singleidentity}
\end{figure}

We first evaluate the accuracy of the neutral mesh reconstructions provided by the IdGen network, as well as the ranked samples according to IdRank. We trained IdGen 
for 400 epochs with a batch size of 128 on an RTX 8000 GPU. The entire training process spans one GPU day. IdRank was trained for 120 epochs with a batch size of 32 on a Titan X GPU. The training process ran for 20 epochs per day, converging over the course of 6 days.
\paragraph*{Baselines and Metrics.}{\label{para:IGN_baseline}}
Since \moniker{} is designed to provide multiple solutions, we establish a baseline by randomly sampling FLAME shape coefficients from its parametric space. We evaluate our approach on the NoW validation benchmark, which allows separate measurement %
for the occluded and unoccluded subsets of the dataset. We compare our method to standard reconstruction techniques as well as state-of-the-art occlusion-based approaches, Diverse3D~\cite{dey2022generating} and FOCUS~\cite{li2023robust}. Our evaluation metrics include the mean, median, and standard deviation of the mean square error (MSE) relative to the ground-truth mesh. %
\paragraph{Results.}
~\cref{fig:qual_Neutral_face_ranking} and~\cref{fig:Neutral_face_singleidentity} present qualitative results of 3D shape reconstruction, %
alongside comparisons with the state-of-the-art neutral face reconstruction method MICA~\cite{zielonka2022mica} and the occluded face reconstruction method by Dey et al.~\cite{dey2022generating}, Diverse3D.
In \cref{fig:qual_Neutral_face_ranking} we observe that, despite mild occlusions, \moniker{} can recover accurate shapes with distinctive features (e.g., cheeks in the first row and the head length in the second row). %
We show an example of extreme occlusion in \cref{fig:Neutral_face_singleidentity}, where we reconstruct two images of the same person wearing a face mask, alongside a minimally occluded reference image.  We observe that \moniker{}, and in particular the top-ranked sample, produces plausible results for the two masked images. 

For quantitative evaluation, we use the NoW benchmark dataset~\cite{Sanyal2019RingNet}, with the results presented in \cref{table:Table_NOW_validation_occlusion_subset_comparison}. Our method achieves a lower average error across all reconstructed samples compared to other state-of-the-art occlusion-based approaches. In addition, the shapes generated by IdGen can achieve competitive results with standard reconstruction methods, as shown by the minimum reconstruction error obtained over all samples (OFER-min).

\paragraph{Ablation.}

\begin{figure}[tbh!]
    \centering
    \includegraphics[width=1\linewidth]{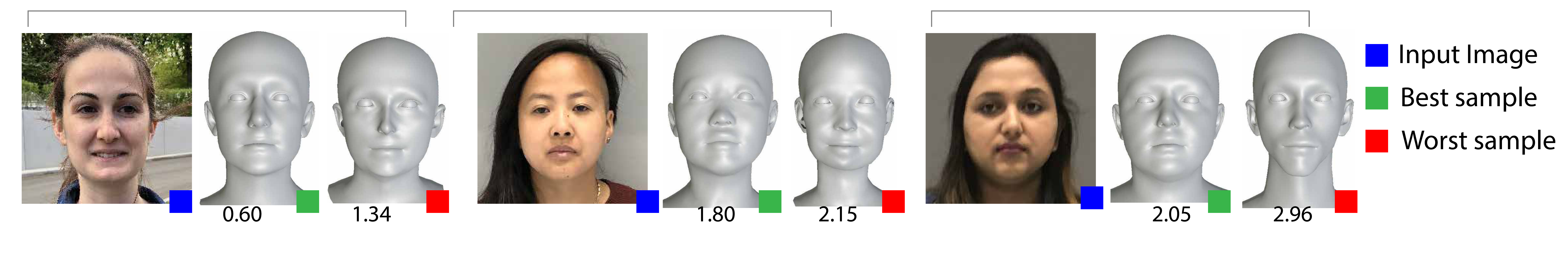}
    \caption{\textbf{Qualitative and quantitative results for best and least ranked samples} for row (b) in \cref{table:Ablation_FLAME_ranking} evaluated on NoW validation images. The median MSE error between the ground truth scan and the reconstructed 3D shape is displayed below each rendering. These results demonstrate that ranking as a selection method enhances the quality of sample selection.}
    \label{fig:Ablation_ranking_qual_diff}
\end{figure}

To highlight the effectiveness of ranking in improving the search space and in selecting high-quality samples, we conducted an ablation study, with results presented ~\cref{table:Ablation_FLAME_ranking}.  %
In row (a), %
we show the results of generating 200 and 1000 random FLAME samples, while row (d) presents the results from \moniker{}. The ``ideal lowest error'' column corresponds to an ideal ranking (i.e. the minimum error within the samples). Even with an increased number of random hypotheses (1000 samples), finding a good reconstruction in this space remains less likely, as indicated by the higher error in the ideal ranking scenario (0.81 with \moniker{} vs 0.90 with FLAME), showcasing the improved search space when ranking is applied to IdGen.
In row (b) %
we show the results of ranking a set consisting of $50\%$ random FLAME samples and $50\%$ \moniker{} reconstructions; row (c) reflects the same for an $80/20$ split. This mimics a scenario where the samples to rank include both correct solutions and high-error ones. Here, the ranked results (1st and 2nd columns) are consistently lower than the average error (3rd column), highlighting its effectiveness in selecting low-error samples. This is further supported by qualitative evidence in~\cref{fig:Ablation_ranking_qual_diff}.

We further demonstrate the network's %
ability to identify subtle differences to select top-ranked sample with lower error than the lower ranked samples, as shown in the color coded error map %
in~\cref{fig:MSE_error_rank_NoW} in the supplement.

\begin{table}
\small
\resizebox{\linewidth}{!}{
\centering
\begin{tabular}{l|c c c|c c c|c c c}
\toprule
 \multirow{2}{*}{Samples} & 
      \multicolumn{3}{c|}{Ideal lowest error Sample} & 
     \multicolumn{3}{c|}{Ranked Sample} & %
     \multicolumn{3}{c}{Average of all Samples}\\
\cline{2-10}
\multirow{2}{*}{} &    
 \small{Med} $\downarrow$ & \small{Mean} $\downarrow$ & \small{std} $\downarrow$ &
 \small{Med} $\downarrow$ & \small{Mean} $\downarrow$ & \small{std} $\downarrow$ &
 \small{Med} $\downarrow$ & \small{Mean} $\downarrow$ & \small{std} $\downarrow$\\
\cline{1-10}
(a)FLAME(200 - 1k)
    & (1.00 - 0.90) & (1.31 - 1.20) & (1.18 - 1.11) 
    & NA & NA & NA
    & 1.75 & 2.19 & 1.86 \\
(b)FLAME(50)+OFER(50)
    & 0.84 & 1.09 & 0.95 
    & 1.08 & 1.44 & 1.32
    & 1.33 & 1.79 & 1.62 \\
(c)FLAME(80)+OFER(20)
   & 0.86 & 1.10 & 0.97
   & 1.34 & 1.81 & 1.65
   & 1.60 & 2.06 & 1.84\\
(d)OFER(100)
    & \textbf{0.81} & \textbf{1.05} & \textbf{0.92} 
    & \textbf{0.98} & \textbf{1.21} & \textbf{1.01}
    & \textbf{1.02} & \textbf{1.25} & \textbf{1.04} \\    

\bottomrule

\end{tabular}
}
\caption{\textbf{Ablation study on the importance of ranking.} \emph{Ideal Lowest Error Sample} refers to the sample with the lowest median MSE. %
\emph{Ranked Sample} is the optimal sample selected by the Ranking Network from all samples generated by the IdGen.
\emph{Average of All Samples} represents the average error across all generated samples. ``NA'' in the Ranked Sample column for the 200-1K range indicates that our network, which was trained on 100 samples, cannot validate these larger sample sizes.\label{table:Ablation_FLAME_ranking}}
\end{table}

\begin{table}
\resizebox{\linewidth}{!}{
\centering
\begin{tabular}{l|c c c|c c c|c c c|}
\toprule
\multirow{3}{*}{Method} & \multicolumn{3}{c|}{Unoccluded} & %
    \multicolumn{3}{c|}{Occluded} & \multicolumn{3}{c|}{Both}\\
\cline{2-10}
\multirow{2}{*}{}
 & \small{Med} & \small{Mean} & \small{std} &
 \small{Med} & \small{Mean} & \small{std} &
 \small{Med} & \small{Mean} & \small{std} \\
\cline{1-10}

FLAME %
    & 1.79 & 2.24 & 1.89 
    & 1.80 & 2.26 & 1.91 
    & 1.79 & 2.25 & 1.90
    \\
\hline

Deep3D~\cite{deng2019accurate}
    & 1.33 & 1.67 & 1.41
    & 1.40 & 1.73 & 1.41
    & 1.36 & 1.70 & 1.41\\
DECA~\cite{feng21deca}
    & 1.18 & 1.47 & 1.24
    & 1.29 & 1.56 & 1.29
    & 1.17 & 1.46 & 1.25\\     %
MICA (4DS)~\cite{zielonka2022mica}
    &  n/a  & n/a & n/a
    & n/a & n/a & n/a
    & 1.02 & 1.25 & 1.05\\
MICA (8DS)~\cite{zielonka2022mica}
    & n/a & n/a & n/a
    & n/a & n/a & n/a
    & 0.90 & 1.11 & 0.92\\
TokenFace~\cite{zhang2023tokenface}
    & n/a & n/a & n/a
    & n/a & n/a & n/a
    & \textbf{0.79} & \textbf{0.99} & \textbf{0.85} \\
\hline
FOCUS~\cite{li2023robust}
    & 1.03 & 1.25 & 1.03
    & 1.07 & 1.34 & 1.19
    & 1.05 & 1.31 & 1.14
    \\
    
FOCUS (MP)~\cite{li2023robust}
    & 1.02 & 1.24 & 1.02
    & 1.08 & 1.34 & 1.20 
    & 1.03 & 1.29 & 1.12
    \\

Diverse3D~\cite{dey2022generating}
    & n/a & n/a & n/a
    & n/a & n/a & n/a
    & 1.41 & 1.78 & 1.52
    \\

OFER-rank (ours)
    & \textbf{0.97}  & \textbf{1.20}& \textbf{1.00} 
    & \textbf{1.01} &\textbf{1.26} &\textbf{1.05} 
    & \textbf{0.98} &\textbf{1.21} &\textbf{1.01} 
    \\

\hline
OFER-min
    & 0.81  &  1.04 & 0.91
    & 0.84 & 1.10 & 0.96
    & 0.81 & 1.05 & 0.92
    \\
    
OFER-avg
    & 1.01  & 1.25 & 1.03
    & 1.04 & 1.29 & 1.08
    & 1.01 & 1.25 & 1.04
    \\

\bottomrule
\end{tabular}
}
\caption{\textbf{Quantitative evaluation on the NoW validation benchmark},
on the unoccluded, occluded and full subsets. 
``FLAME'' 
is created by sampling random FLAME shape coefficients. %
Rows 2-6 show standard reconstruction methods, while the bottom rows show occlusion-based approaches.
OFER-rank is our result using IdGen and IdRank networks. 
The bottom two rows show the minimum (OFER-min) and average (OFER-avg) reconstruction errors obtained over the 100 samples provided by IdGen.
We show median (med), mean, and standard deviation (std) of the non-metrical 
MSE between the reconstructed and ground truth shapes.}
\label{table:Table_NOW_validation_occlusion_subset_comparison} 
\end{table}

\subsection{Evaluation on reconstructions with ExpGen}
\label{ssec:results_exp}

We run experiments on the full expressive reconstructions enabled by ExpGen. 
Similarly to IdGen, the network was trained for 400 epochs with a batch size of 128 on an RTX 8000 GPU, with the training spanning one GPU day.

\begin{figure}
    \centering
    \includegraphics[width=1.0\linewidth]{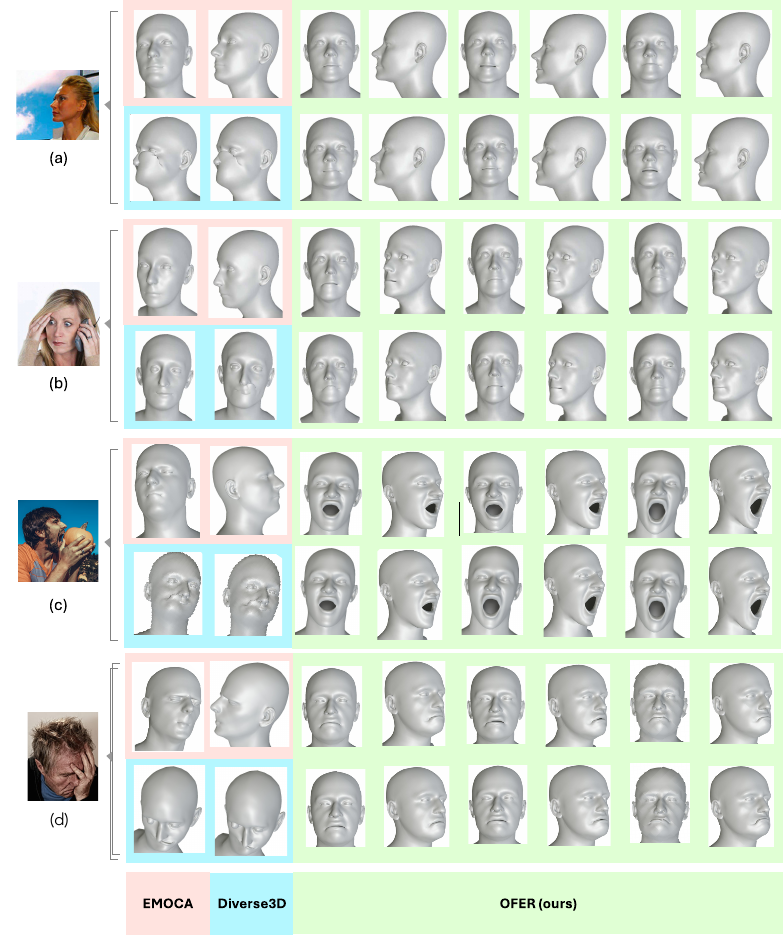}
     \caption{\textbf{Comparison of expression reconstruction for in-the-wild occluded images.} We compare against EMOCA~\cite{danecek2022emoca} (showing front and side view, pink), two reconstructions from Diverse3D~\cite{dey2022generating} (blue), and six samples (front and side view) from our method (green). }
    \label{fig:aflw2000}
\end{figure}

\begin{table}
\resizebox{\linewidth}{!}{
\centering
\begin{tabular}{l| c c c c c}
\toprule
\multirow{2}{*}{Method} & %
    \multicolumn{1}{c|}{$N_C^{unocc}$} &
    \multicolumn{1}{c|}{$N_C^{occ}$} &
    \multicolumn{1}{c|}{$E_D^{occ}$} &  
    \multicolumn{1}{c|}{$E_C^{unocc}$} &  
    \multicolumn{1}{c|}{$E_C^{occ}$}\\
\cline{2-6}
\multirow{2}{*}{}
 &  \small{RMSE} $\downarrow$ & 
  \small{RMSE} $\downarrow$ & 
  \small{RMSE} $\downarrow$ & 
 \small{RMSE} $\downarrow$ & 
 \small{RMSE} $\downarrow$  \\
\cline{1-6}

Diverse3D\cite{dey2022generating}
      & 2.44 & 2.16 & 6.4 & 7.46 & 4.72 \\
OFER (ours)
     & 1.90 & 1.95 & \textbf{3.2} & \textbf{3.48} & \textbf{2.78}\\

\bottomrule
\end{tabular}
}
\caption{\textbf{Quantitative evaluation using SE}
on unoccluded ($unocc$) and occluded images ($occ$) in Dey and CO-545 datasets. The letters represent: $N$ for Neutral, $E$ for expression, %
$D$ for Dey dataset, 
and $C$ for CO-545 dataset. Reconstructions evaluated on all front face vertices for $unocc$ images and only on unoccluded vertices for $occ$ images.} %
\label{table:Table_COMA_validation_comparison} 
\end{table}

\begin{table}
\resizebox{\linewidth}{!}{
\centering
\begin{tabular}{l| c c c c c c c}
\toprule
\multirow{2}{*}{Method} & %
    \multicolumn{1}{c|}
    {$\text{CSE}$} &  
    \multicolumn{1}{c|}{$\text{minASD-V}$} & 
    \multicolumn{1}{c|}{$\text{maxASD-V}$} & 
    \multicolumn{1}{c|} 
    {$\text{ASD-O}$} & 
    \multicolumn{1}{c|}
    {$\text{ODE}$} &
    \multicolumn{1}{c|}
    {$\text{ODE-O}$} & \multicolumn{1}{c|}
    {$\text{ODE-V}$}\\
\cline{2-8}
\multirow{2}{*}{}
 &  \small{RMSE} $\downarrow$ & 
  \small{RMSE} $\downarrow$ & 
  \small{RMSE} $\downarrow$ & 
  \small{HM} $\uparrow$ & 
 \small{max md} $\downarrow$ & 
  \small{max md} $\downarrow$ & 
 \small{max md} $\downarrow$ \\ 
\cline{1-8}

Diverse3D\cite{dey2022generating}
      & 0.30 & 1.31 & 1.96 & 0.16 & 0.4 & 0.57 & 0.33\\
OFER (ours)
     & \textbf{0.17} & \textbf{0.65} & \textbf{1.39} & 
     \textbf{0.18} &
     \textbf{0.01} & 
     \textbf{0.01} & \textbf{0.01}\\

\bottomrule
\end{tabular}
}
\caption{\textbf{Quantitative evaluation of expression reconstructions in terms of CSE, ASD-V and ODE} evaluated using CO-545 dataset. The metrics are details in ~\cref{para:expevalmetric}}
\label{table:Table_CSE_ASD_metric} 
\end{table}

\begin{table}
\resizebox{\linewidth}{!}{
\centering
\begin{tabular}{l| c c c c c c}
\toprule
\multirow{3}{*}{Method}
& %
    \multicolumn{3}{c|}{$\text{STD-S}$} &
    \multicolumn{3}{c|}{$\text{ODE}$} \\
\cline{2-7}
& %
    \multicolumn{1}{c|}{$mask$} &
    \multicolumn{1}{c|}{$sunglasses$} &
    \multicolumn{1}{c|}{$naturalocc$} &
    \multicolumn{1}{c|}{$mask$} &
    \multicolumn{1}{c|}{$sunglasses$} &
    \multicolumn{1}{c|}{$naturalocc$} \\
\cline{2-7}
\multirow{2}{*}{}
 &  \small{RMSE} $\uparrow$ 
  &  \small{RMSE} $\uparrow$ 
   &  \small{RMSE} $\uparrow$  
    &  \small{RMSE} $\downarrow$ 
  &  \small{RMSE} $\downarrow$  
   &  \small{RMSE} $\downarrow$  \\ 
\cline{1-7}

Diverse3D\cite{dey2022generating}
      & 11.81 & 21.28 & 20.84 & 0.95 & 0.34 & 0.41\\
OFER (ours)
     & \textbf{34.04} & \textbf{34.38} & \textbf{34.56} & \textbf{0.002} & \textbf{0.001} & \textbf{0.003}\\

\bottomrule
\end{tabular}
}
\caption{\textbf{Quantitative evaluation of expression reconstructions using STD-S and ODE metrics} evaluated on 200 images per subset (mask, sunglasses and natural occlusions) from ~\cite{erakiotan2021recognizing}.}

\label{table:Table_std_metric} 
\end{table}

\paragraph{Baselines and Metrics.}{\label{para:expevalmetric}}
We compare our method against the state-of-the-art diverse expression reconstruction approach by Dey et al.~\cite{dey2022generating} and the single-hypothesis expression reconstruction method EMOCA~\cite{danecek2022emoca}. It is important to note that Dey et al.~\cite{dey2022generating} employs test-time optimization, whereas our approach does not. %
Additionally, their network is trained on the COMA dataset, the same dataset used in the creation of CO-545. As a result, the evaluation data overlaps with their training set, potentially giving their results an additional advantage. %

Building on the metrics introduced by ~\cite{yuanDDP2020} and ~\cite{dey2022generating}
we evaluate on six metrics: (1) Sample Error (\textbf{SE}): %
the average Root Mean Square Error (RMSE) between the landmark vertices of the reconstructed samples and the ground truth; (2) Closest Sample Error (\textbf{CSE}): %
the SE of the closest reconstructed sample; 
(3) Average Self Distance-Visible (\textbf{ASD-V}): to ensure that the visible regions of the reconstructions remain close to the input, the maximum distance across samples should be minimized. \textbf{maxASD-V} measures the average per-vertex RMSE on the visible regions of a sample and its ``farthest'' neighboring sample across all instances, while \textbf{minASD-V} measures ASD-V with the ``nearest'' neighbour sample; %
(4) Average Self Distance-Occluded (\textbf{ASD-O}): the reconstructions of \emph{occluded} regions should show high diversity, given by $d^o = \frac{\text{RMSE(gt}^v{\text{,x}_i^v}{\text{)}}}{\text{max(MD)}^o}$, while the visible regions must remain close to the ground truth, given by $d^v = \frac{\text{RMSE(x}_i^o{\text{,x}_j^o}{\text{)}}}{\sqrt{\text{max(MD)}^o}}$. Here, $\text{max}(MD^o)$ is the maximum mahalanobis distance threshold of the occluded vertices, and $x_j$ is the nearest neighbour sample of $x_i$ based on the L2 distance of occluded vertices. The superscript $v$ and $o$ point to the visible and occluded vertices. We compute ASD-O as the harmonic mean (HM) between $d^v$ and $d^o$,  %
given by %
~\cref{eq:asd-o}; 
(5) Shape Standard Deviation (\textbf{STD-S}): in case of absence of ground-truth 3D data, we opt for the standard deviation metric of the entire shape, excluding data which falls outside the maximum mahalanobis distance per vertex calculated using COMA~\cite{ranjan2018coma}; 
(6) Out-of-Distribution Error (\textbf{ODE}): ODE measures the average per-vertex mahalanobis distance (MD) out-of-distribution error.  %
It addresses the observation that reconstructions from Dey et al. \cite{dey2022generating} occasionally produce non-plausible expressions that %
do not fall within the distribution of ground truth expressions. %
 \begin{align}
   &\text{ASD-O} = \frac{2 \cdot (1 - d^v) \cdot d^o} {(1 - d^v) + d^o}\label{eq:asd-o}
  \end{align}

\paragraph{Results.}
The qualitative comparison of diverse expressive faces for occluded images is shown in~\cref{fig:exp_comparison} and~\cref{fig:aflw2000}. Our approach effectively generates plausible faces with varied expressions, even under challenging occlusions such as face masks. In contrast, EMOCA~\cite{danecek2022emoca} tends to produce degenerate faces in similar hard occlusion scenarios, while Dey et al.~\cite{dey2022generating} often results in unrealistic and extreme reconstructions. The quantitative evaluation shown in~\cref{table:Table_COMA_validation_comparison}, ~\cref{table:Table_CSE_ASD_metric} and ~\cref{table:Table_std_metric} also supports this, particularly highlighting the substantial improvement over Dey et al.~\cite{dey2022generating}. We sampled 15 expressions per image for this evaluation. Our method demonstrates a significant advantage, reliably reconstructing visible regions and producing varied expressions in occluded areas.

\section{Limitations and Future work}
\label{sec:limitations}
3D face reconstruction offers many valuable applications, but if not carefully regulated, it can lead to negative societal impacts, including privacy violations and misuse in surveillance, such as deepfakes or identity theft. To address these concerns, we ensured that the data we used was anonymized and that we utilized datasets that were ethically sourced.

The generative capabilities of diffusion models with diverse inputs are attributed to the amount and variations in the training data~\cite{samuel2024OOD}. In our case, we had access to only a small subset of the available 3D supervision dataset for both IdGen and ExpGen, As a result, our generated expressions may lack certain expressive details. %
Therefore, a promising direction for future research is to integrate both 2D and 3D supervision to improve the performance of the method. 

When we refer to performance, we specifically measure it in terms of how closely the reconstructed model approximates the ground-truth scan. Although our ranking network selects a sample that minimizes the error compared to a randomly chosen one from the generated set, it does not guarantee the selection of the absolute best match, i.e., the sample with the lowest error and closest approximation to the ground truth. 
From this observation we identify two potential research directions for constructing an ideal ranking network. The first is to refine the ranking architecture itself by adopting a more suitable loss function. Softmax scoring is not ideal when presented with large number of sample sets %
since it dilutes the scores, making it challenging to identify the best among the higher-quality samples. Additionally a more sophisticated ranking mechanism~\cite{Brito2014PointVsListRank} can complement to the selection of quality samples. A second promising approach involves integrating the ranking process as a feedback mechanism during training of the diffusion model. Rather than using a stand-alone model, this integration could improve the quality of the generated samples, resulting in improved overall performance.

\section{Conclusion}
\label{sec:conclusion}
In this paper we introduced OFER, a conditional diffusion-based method for generating multiple hypotheses of expressive faces from a single-view, in-the-wild occluded image. Key to OFER is the use of two diffusion models to generate FLAME 3DMM shape and expression coefficients. To ensure a consistent geometric face shape for varied expression of a single identity, we introduced a probabilistic ranking method to select an optimal sample from the generated shape coefficients. By combining the statistical learning strengths and generative capabilities of diffusion models, along with the smooth face reconstruction provided by a parametric model, our method produces plausible 3D faces that accurately reflect the input image. OFER achieves state-of-the-art results in diverse expression reconstruction outperforming existing occlusion-based methods, and can generate plausible and diverse results for a given input.

{
    \balance
    \small
    \bibliographystyle{ieeenat_fullname}
    \bibliography{main}
}
\clearpage
\setcounter{page}{1}
\maketitlesupplementary
\label{sec:supplement}

\section{Ablation Study}
We conducted several ablation studies on the choice of conditioning signal, embedding, and loss functions to optimize our model networks, which are detailed in the following sections.

\subsection{Identity Generative Network (IdGen)}
\paragraph{Choice of image embedding.}
The representation used for the conditioning image provided to the diffusion network plays an important role in the reconstruction task. In \cref{table:ablation_embedding} we show ablation results for the embedding used in IdGen. The results indicate that using only the ArcFace~\cite{Deng2018ArcFaceAA} embedding improves the error metric of this network, compared to employing FaRL~\cite{zheng2022general} or a combination of both as conditions. This may be because ArcFace~\cite{Deng2018ArcFaceAA} is used to distinguish identity-defining features, while FaRL~\cite{zheng2022general} is designed for downstream facial analysis tasks that require capturing facial subtleties. In addition, we compare the performance of ArcFace with state of the art image embeddings, DINOv2~\cite{oquab2024dinov} and CLIP~\cite{radford2021clip}. The results are presented in~\cref{tab:idemb_ablation}. The reason we hypothesize for the lower performance of CLIP and DINOv2 is that they are trained on multi-domain models and not specifically trained on faces like ArcFace.

\subsection{Identity Ranking Network (IdRank)}
\begin{figure}[hbt!]
    \centering
    \includegraphics[width=1.0\linewidth]{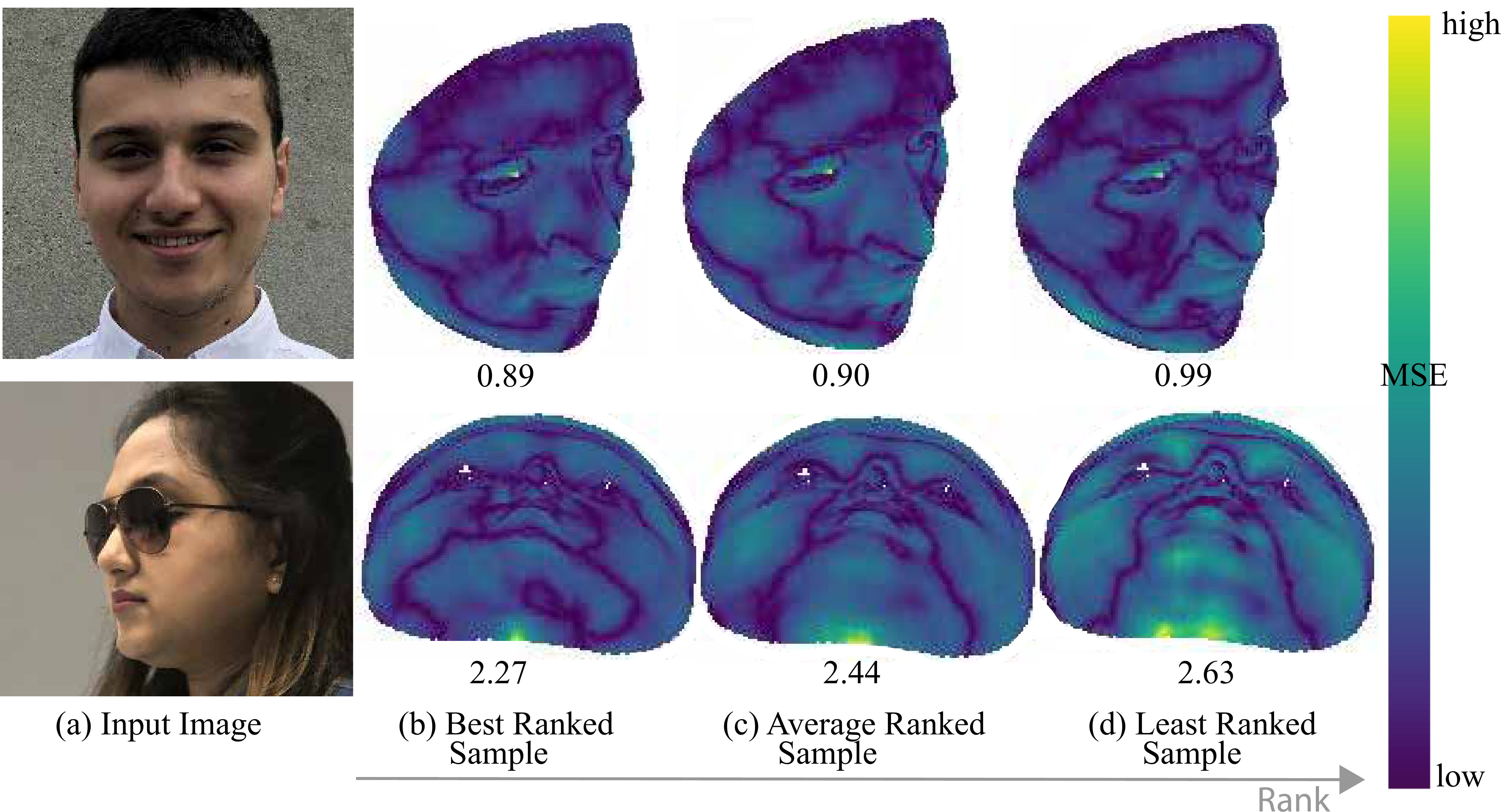}
   \caption{\textbf{Ranking on NoW validation images}. For the image in column (a), we generate 100 shape coefficients with {IdGen} and rank them using {IdRank}. %
   Column (b) shows the top-ranked sample, (c) a mid-ranked sample (50th), and (d) the worst-ranked sample. Columns (b)-(d) show viewpoints highlighting the sample errors, with its median MSE shown underneath each sample.} %
    \label{fig:MSE_error_rank_NoW}
\end{figure}

\paragraph{Ranking selects low-error samples}
We showcase the ability of ranking network to identify subtle differences to select top-ranked sample with lower MSE error than the lower ranked samples in~\cref{fig:MSE_error_rank_NoW}.

\begin{table}
\resizebox{\linewidth}{!}{
\centering
\begin{tabular}{l|c c c| c c c}
\toprule
\multirow{3}{*}{Embedding} & \multicolumn{3}{c|}{Occluded subset} &
\multicolumn{3}{c|}{Unoccluded subset} \\
    
\cline{2-7}
\multirow{2}{*}{}
 & \small{Med} $\downarrow$ & \small{Mean} $\downarrow$ & \small{std} $\downarrow$ 
 & \small{Med} $\downarrow$ & \small{Mean} $\downarrow$ & \small{std} $\downarrow$ \\
\cline{1-7}

ArcFace~\cite{Deng2018ArcFaceAA}
    & \textbf{1.03} & \bf{1.26} & \bf{1.06} & \bf{1.01} & \bf{1.22} & \bf{1.01}\\
\hline
ArcFace+FaRL 
    & 1.05 & 1.30 & 1.09 & 1.03& 1.28& 1.06\\
\hline
FaRL~\cite{zheng2022general}
    & 1.15 & 1.40 & 1.12 & 1.07 & 1.30 & 1.06 \\
\bottomrule
\end{tabular}
}
\caption{\textbf{Ablation on %
image embeddings for the IdGen network}. %
MSE error %
on the NoW validation dataset, using %
ArcFace~\cite{Deng2018ArcFaceAA} 
embedding, FaRL~\cite{zheng2022general} embedding, or a combination of both.
}
\label{table:ablation_embedding} 
\end{table}

\begin{table}
    \centering
    \begin{tabular}{l|c c c}
         Method & Arcface & CLIP & DINOv2 \\
         \hline
         Median & 1.01 & 1.65 & 1.68\\
         Mean & 1.24 & 2.07 & 2.08\\
    \end{tabular}
    \caption{\textbf{Ablation study for IdGen embedding} on NoW validation benchmark. ArcFace trained for identity clustering performes better in comparison to other image embedding for identity recognition reconstruction task.}
    \vspace{-4mm}
  \label{tab:idemb_ablation}
\end{table}  

\paragraph{Sampling Quantity for Input.}
Naturally, increasing the number of generated samples results in a more thorough coverage of the data distribution. %
As the sample set size increases and leads to denser mode coverage, we expect the average error to get closer to the ground truth. However, for the ranking network to identify that optimal element, it must be be trained on a sufficiently large number of samples to effectively rank them and select the best one. This requirement comes with a significant computational costs, %
especially due to the slower inference time of diffusion models. Thus, finding the right balance in the number of samples used for training is important. We conducted an ablation study on different sample sizes generated by IdGen, and assessed the median error of the optimal sample from each set to determine the ideal number of samples for training IdRank. The results are presented in ~\cref{tab:Ablation_lowest_mse_samples}.

\begin{table}
\centering
\begin{tabular}{l|c c c}
\toprule
\multirow{3}{*} {}\#samples & %
    \multicolumn{3}{c}{Ideal lowest error sample} \\
\cline{2-4}
\multirow{2}{*}{}
  & \small{Med} $\downarrow$ & \small{Mean} $\downarrow$ & \small{std} $\downarrow$ \\
\cline{1-4}
FLAME (Baseline) & 1.02 & 1.30 & 1.11 \\
\cline{1-4}
10
    & 0.89 & 1.13 & 0.97  \\
50
   & 0.83 & 1.07 & 0.94\\
100
   & 0.81 & 1.05 & 0.92 \\
500
   & 0.78 & 1.02 & 0.90 \\
\bottomrule
\end{tabular}
\caption{\textbf{Ablation on effect of sampling quantity as input to IdRank.} To evaluate the tradeoff between the number of generated samples by IdGen and the "ideal" lowest error sample in the set. The values are the lowest median error sample evaluated against the Now validation benchmark.}%
\label{tab:Ablation_lowest_mse_samples}
\end{table}

\paragraph{Face vertices as Input.}
Selecting the most accurate sample requires a network design that excludes irrelevant information (See~\cref{para:problem_setting} for more details). To validate this claim, we conducted an ablation study comparing three inputs: (a) shape coefficients, which are a PCA model representing entire head shape; (b) all landmark vertices of the reconstructed head shape; and (c) front-face vertices from the reconstructed head shape. The results of this comparison are presented in \cref{table:ablation_rank_input}. The high errors associated with the coefficients and vertex-based reconstructions likely stem from the inclusion of irrelevant information.

\paragraph{Loss function.}
The challenge in selecting the best representative shape from the samples generated by the IdGen is that, even with mild occlusions where most of the facial structure is preserved, variability can still occur. This means we need an appropriate loss function that mitigates the selection of sub-optimal samples. We considered two loss functions for this network: binary cross-entropy and softmax loss, which approximates the ground truth error distribution considering all samples. To assess the performance of both, we conducted an ablation study, the results of which are presented in ~\cref{table:loss_selection}. Since more than one sample can be optimal, ranking and selection using softmax loss yields better precision with a set of higher-ranked samples.
\begin{table}
\resizebox{\linewidth}{!}{
\centering
\begin{tabular}{l|c c c|c c c|c c c}
\toprule
\multirow{2}{*}{Input to rank network} & \multicolumn{3}{c|}{precision $\uparrow$} & %
    \multicolumn{3}{c|}
{IOU $\uparrow$} &
    \multicolumn{3}{c|}
{error (GT/Pred) (ideal = 1) } \\
\cline{2-10}
\multirow{2}{*}{}
 & \small{1} \% & \small{10} \% & \small{20} \% &
 \small{10} \% & \small{20} \% & \small{30} \% &
 avg1 & avg5 & avg10 \\
\cline{1-10}
(a) $\hat{S}$
     & 3.3 & 21.3 & 33.2
     & 26.7 & 40.0 & 53.3
     & 0.54 & 0.57 & 0.64 \\
(b) $M^{\text{frontal}}$
    & 3.3  & 11.3 & 24.8
    & 13.3 & 30.0 & 53.3 
    & 0.49 & 0.58 & 0.61\\
(c) X ({\tiny{$\mu_{M},M'$}})
    & 3.3 & \textbf{44.5} & \textbf{64.7}
    & \textbf{33.3} & \textbf{80.0} & \textbf{93.3}
    & \textbf{0.68} & \textbf{0.78} & \textbf{0.83}\\
\bottomrule
\end{tabular}
}
\caption{\textbf{Ablation for input to ranking network.} $\hat{S}$ is the 300-dimentional shape coefficients generated from IdGen; $M^{\text{frontal}}$ is the front face vertices of FLAME mesh $M$ reconstructed from ($\hat{S}$ ; X=($\mu_M, M'$) is the mean, residual pair defined in ~\cref{eq:mean_residual}}
\label{table:ablation_rank_input} 
\end{table}

\begin{table}
\resizebox{\linewidth}{!}{
\centering
\begin{tabular}{l|c c c|c c c|c c c}
\toprule
\multirow{2}{*}{Loss} & \multicolumn{3}{c|}{precision $\uparrow$} & %
    \multicolumn{3}{c|}
{IOU$\uparrow$} &
    \multicolumn{3}{c|}
{error (GT/Pred)$\downarrow$} \\
\cline{2-10}
\multirow{2}{*}{}
 & \small{1} \% & \small{10} \% & \small{20} \% &
 \small{10} \% & \small{20} \% & \small{30} \% &
 avg1 & avg5 & avg10 \\
\cline{1-10}

BCE loss 
    & 0.0 & 3.0 & 20.8 
    & 0.0 & 0.0 & 16.7 
    & 0.56/0.86 & 0.60/0.88 & 0.63/0.85 \\
\hline
Softmax loss 
    & 0.0 & \textbf{30.0} & \textbf{51.7}
    & \textbf{50.0} & \textbf{66.7} & \textbf{66.7} 
    & \textbf{0.58/0.76} & \textbf{0.62/0.77} & \textbf{0.65/0.79} \\
\bottomrule
\end{tabular}
}
\caption{\textbf{Ablation for loss function for IdRank.} We trained the network with Binary Cross Entropy (BCE) loss and Cross Entropy on Softmax (Softmax) loss using 100 Stirling(HQ)~\cite{feng2018evaluation} frontal face images. For validation, we used 20 Florence~\cite{bagdanov2011florence} dataset frontal face images. IOU represents intersection over union of predicted ranking order and ground truth ranking order for first 10, 20 and 30 sorted rank index. error(GT/Pred) shows the average error for first 1, 5 and 10 ground truth rank samples and that of predicted rank samples.}
\label{table:loss_selection} 
\end{table}

\begin{figure}[hbt!]
    \centering
    \includegraphics[width=1\linewidth]{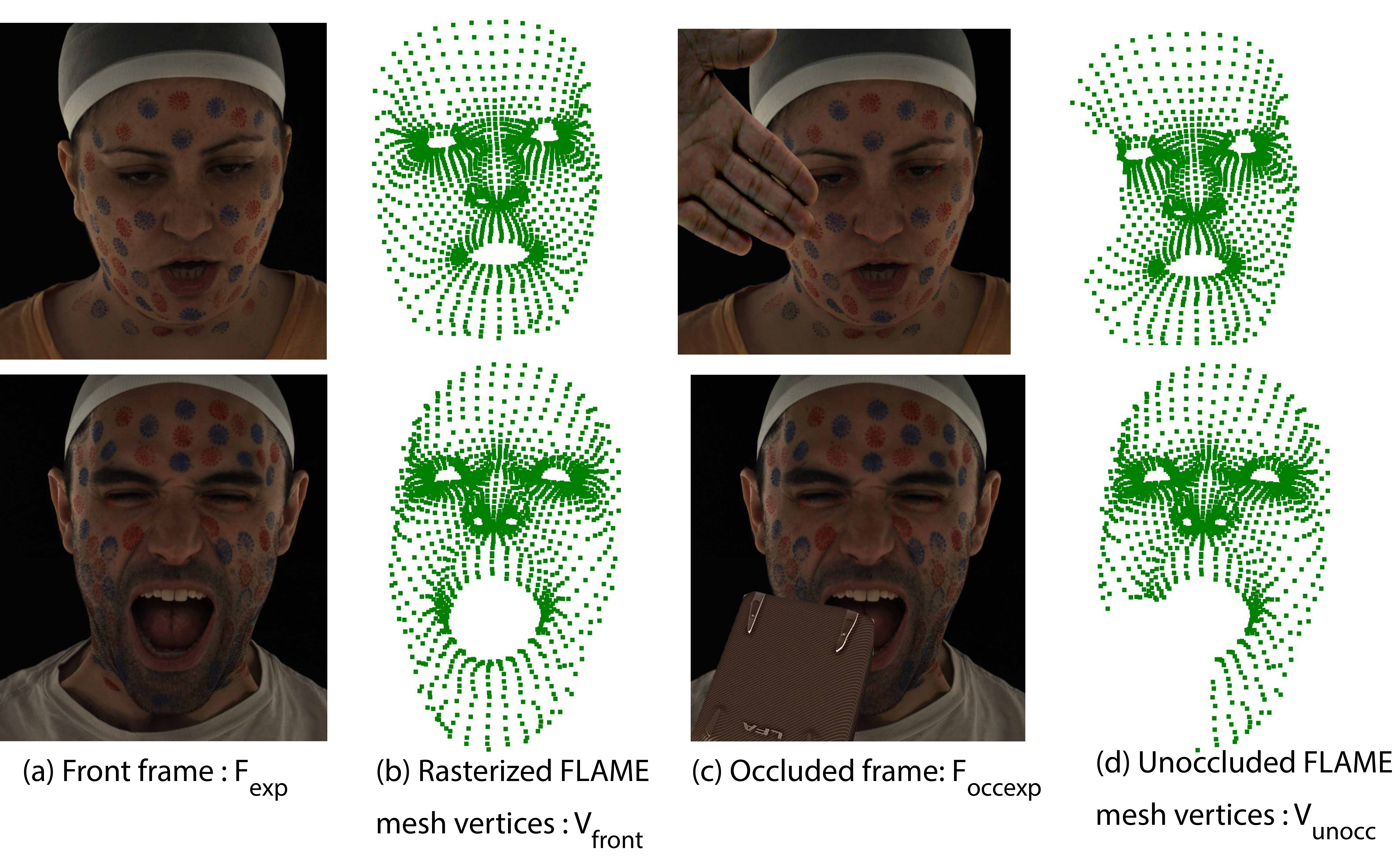}
    \caption{\textbf{CO-545.} Column (a)  the expressive frames with frontal view, $F_{exp}$; Column (b) the rasterized the mesh vertices $V_{front}$; Column (c) the occluded frames with synthetic objects,$F_{occexp}$; Column (d) the unoccluded vertices which belong to unoccluded pixels, $V_{unocc}$. The pairs ($F_{occexp}, V_{unocc}$) make CO-545 dataset.}
    \label{fig:co-545}
\end{figure}

\section{Design Choices}

\subsection{Ranking by distribution matching.}
The objective of the ranking model is to sort the output from the network and optimize a loss based on ranking order. We choose a list-wise ranking method %
due to its demonstrated effectiveness~\cite{Brito2014PointVsListRank}.
Since sorting and ranking are non-differentiable operations, we reformulate the sort-rank problem into a probability distribution alignment problem aiming to minimize the softmax loss $L_R$ between the ground truth distribution $g$ and the predicted distribution $h$. 

\subsection{Exclusion of Ranking in Expression Generation}
Our ranking network is trained to rank only \emph{neutral} shapes %
and not expressions. This is because shape geometry remains consistent despite occlusions or variations in expressions in the input image. But, ranking the expression hypotheses is harder since multiple hypotheses can be equally valid for occluded regions.

\section{Architecture}
\begin{figure}
    \centering
    \includegraphics[width=1.0\linewidth]{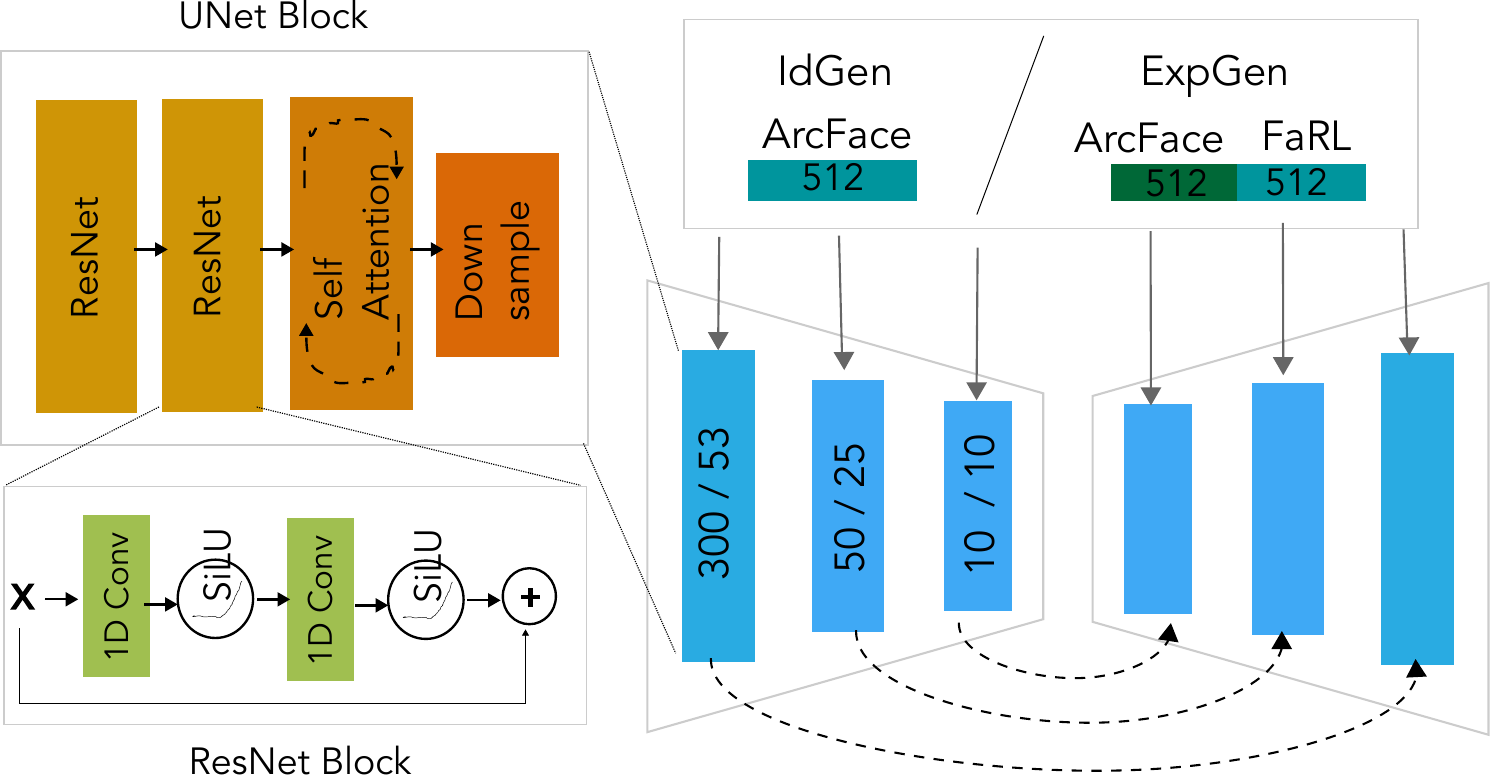}
    \caption{\textbf{1D Unet Transformer Hybrid architecture} of IdGen and ExpGen. Each block of the UNet is comprised of two resnet blocks, followed by a self-attention module and a downsampling/upsampling module for the encoder and decoder respectively. Each resnet block consists of two 1D convnet followed by SiLU activation and residual connection. For IdGen, the conditional embedding ($\real{512}$) is passed to every layer of the UNet. The training input to IdGen is FLAME~\cite{li2017flame} shape coefficient, $S \in \real{300}$ which gets downsampled to $50$ and to $10$ at the bottleneck. For ExpGen, the conditional embedding is ($\real{1024}$) and the input is FLAME expression (including 3 jaw cofficients) $E \in \real{53}$ which is downsampled to $25$ and to $10$ at the bottleneck.}

    \label{fig:1d_unet}
\end{figure}

The detailed overview of our 1-D Unet-transformer hybrid architecture of our IdGen and ExpGen networks is shown in~\cref{fig:1d_unet}. Both share similar architecture, differing in the embeddings and the inputs. The conditional embedding of IdGen obtained from ArcFace~\cite{Deng2018ArcFaceAA} is $\real{512}$, and the embedding of ExpGen obtained by concatenating ArcFace~\cite{Deng2018ArcFaceAA} and FaRL~\cite{zheng2022general} embeddings is $\real{1024}$. 

\section{Datasets}
\subsection{Training Dataset}
In~\cref{table:Dataset_without_mica}, we list the datasets used to train the networks in our framework. FaMoS~\cite{bolkart2023Famos} dataset comprises of 3D registered FLAME meshes. %
We utilized only a subset of them such that it covered all the expression variations in the entire dataset. Please note that our network trains in parametric space rather than on meshes. Therefore, we obtained the corresponding parameters for this subset directly from the authors of FaMoS. We used the expression coefficients from this subset to train the ExpGen. The remaining four datasets listed in the table (Stirling~\cite{feng2018evaluation}, Florence~\cite{bagdanov2011florence}, Facewarehouse~\cite{cao2013facewarehouse}, LYHM~\cite{hang2020lyhm}) were used to train IdGen and IdRank.
\begin{table}
\centering %
\begin{tabular}{l| c c c}
\toprule
\multirow{2}{*}{ Dataset} &
\multirow{2}{*} {}Num & Num & with\\
& Subjects & Images & exp \\
\midrule
Stirling~\cite{feng2018evaluation} &  133 & 1322 & \xmark \\
Florence~\cite{bagdanov2011florence} &  53 & 1239 &\xmark \\
Facewarehouse~\cite{cao2013facewarehouse} &  150 & 3000 &\xmark \\
LYHM~\cite{hang2020lyhm} &  1211 & 7118 &\xmark \\
\midrule
FaMoS~\cite{bolkart2023Famos} &  95 & ~1.5M &\cmark \\
\bottomrule
\end{tabular}
\caption{\textbf{Datasets used for training Networks}. FaMoS expression coefficients are used to train ExpGen. The shape coefficients from the remaining four datasets were used to train IdGen and IdRank.}
\label{table:Dataset_without_mica}
\end{table}
\subsection{CO-545 Evaluation Dataset}
We introduce a new dataset named CO-545 to quantitatively evaluate occluded expressions. First, we select the middle frame of each sequence in the COMA dataset, $F_{exp}$, which exhibits frontal views with expressive features while excluding neutral expressions. Subsequently, we rasterize the FLAME mesh for each frame to eliminate naturally occluded vertices from the camera's perspective, selecting only facial vertices and excluding those from the back of the head, eyeballs, neck, and ears. This subset of vertices is denoted as $V_{front}$. Occlusion masks~\cite{Voo2022synocc} are then applied to the selected frames $F_{occexp}$, removing additional vertices from $V_{front}$ that fall within the masked pixel areas of the image. We thus obtain the set of unoccluded vertices, denoted as $V_{unocc}$, for each masked image. The ($F_{occexp}, V_{unocc}$) pairs form the dataset, allowing us to evaluate occluded samples only within the visible regions. This procedure enables the inclusion of additional evaluation data to the dataset. A few samples from this dataset are shown in~\cref{fig:co-545}. 

\section{Additional Results.}
\subsection{Quantitative Results}
We provide comprehensive comparison of neutral face reconstruction, including additional methods specifically focused on this task, alongside the occlusion-based reconstruction methods discussed in the main paper. %
~\cref{table:Table_NOW_validation_occlusion_subset_comparison} presents the evaluation results for non-metrical reconstruction error on the NoW~\cite{Sanyal2019RingNet} validation benchmark, while~\cref{table:now_test} presents metrical reconstruction error on the NoW test benchmark. We note that our method does not outperform TokenFace~\cite{zhang2023tokenface}, which is explicitly trained with both 2D and 3D supervision--a limitation acknowledged and addressed as future work in the main paper. However, in the case of MICA, which when trained only on the four datasets, the results in \cref{table:Table_NOW_validation_occlusion_subset_comparison} show that its performance is similar to our method. In addition, OFER demonstrates improved results when ranking is incorporated.
\begin{table}
\resizebox{\linewidth}{!}{
\centering %
\begin{tabular}{l| c c c}
\toprule
\multirow{2}{*}{ Method} &
\multirow{2}{*} {}Median $\downarrow$ & Mean $\downarrow$ & Std  $\downarrow$ \\
  & (mm) & (mm) & (mm) \\
\midrule
TokenFace \cite{zhang2023tokenface} & \textbf{0.97} & \textbf{1.24} & \textbf{1.07} \\
MICA (8DS) \cite{zielonka2022mica}  &  1.08 & 1.37 & 1.17 \\

3DDFA V2 & 1.53 & 2.06 & 1.95 \\
DECA \cite{feng21deca} & 1.35 & 1.80 & 1.64 \\
Dib et al. \cite{dib2021towards} & 1.59 & 2.12 & 1.93\\
Dense landmarks \cite{wood20223d} & 1.36 & 1.73 & 1.47\\
FOCUS-MP \cite{li2023robust} & 1.41 & 1.85 & 1.70\\
Deng et al \cite{deng2019accurate} & 1.62 & 2.21 & 2.08\\
RingNet\cite{Sanyal2019RingNet} & 1.50 & 1.98 & 1.77 \\
OFER (4DS) (sample selected by ranking) & 1.27 & 1.64 & 1.29 \\

\bottomrule
\end{tabular}
}
\caption{\textbf{Neutral face 3D Metrical reconstruction error on the NoW test benchmark.}. The results show comparison of accuracy of single view reconstruction methods based on NoW challenge.} 
\label{table:now_test}

\end{table}

\subsection{Qualitative Results.}
\begin{figure*}
    \centering
    \includegraphics[width=0.8\linewidth]{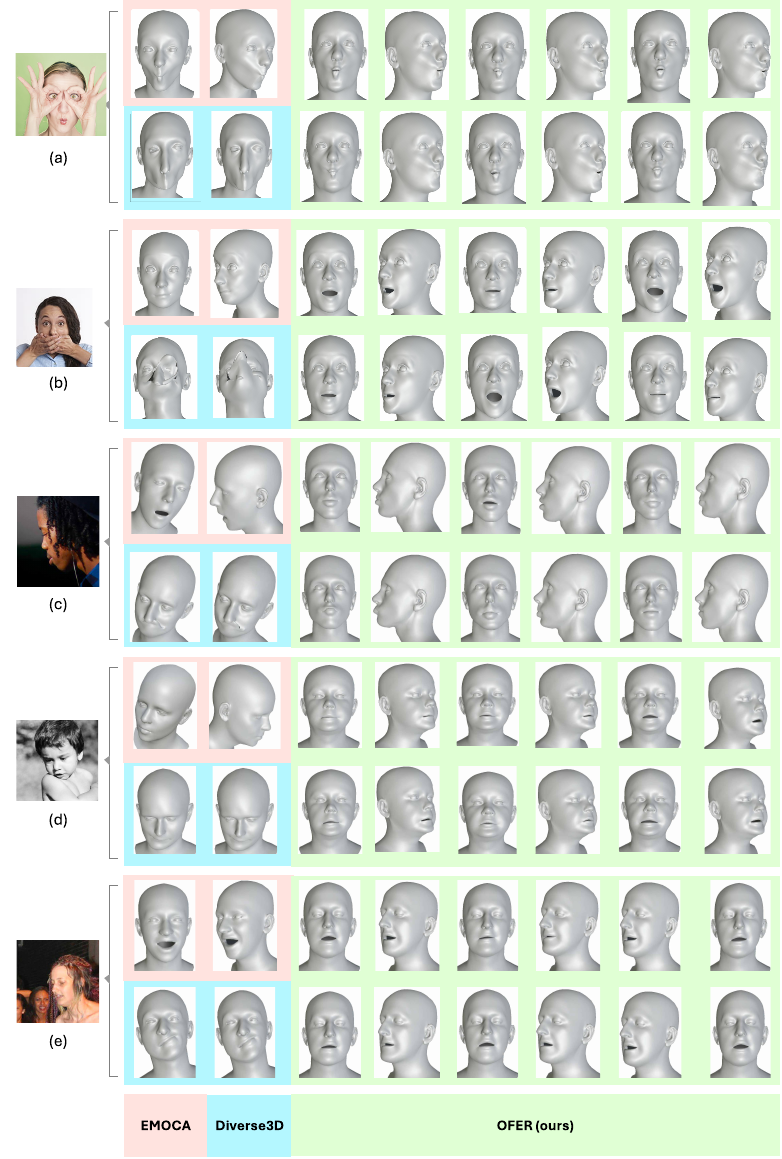}
     \caption{\textbf{Comparison of expression reconstruction for in-the-wild occluded images.} We compare against EMOCA~\cite{danecek2022emoca} (showing front and side view, pink), two reconstructions from Diverse3D~\cite{dey2022generating} (blue), and six samples (front and side view) from our method (green). }
    \label{fig:aflw2000_all}
\end{figure*}

\begin{figure}[tbh!]
    \centering \includegraphics[width=1\linewidth]{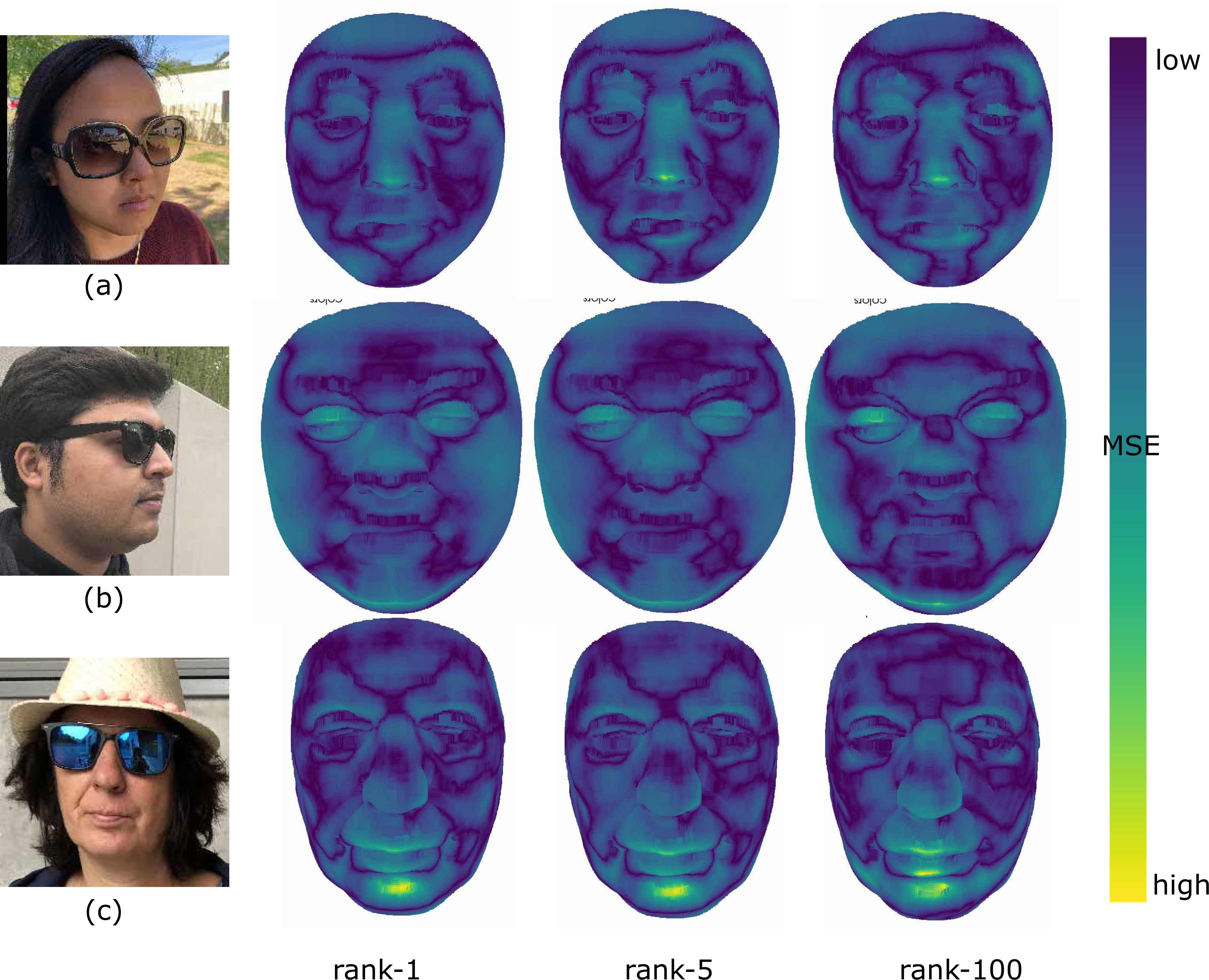}
    \caption{\textbf{Ranking on NoW validation}. Column 1 shows the optimal sample selected by IdRank, column 2 displays the sample ranked 5th, and the last column shows the lowest-ranked sample. Although error differences are subtle, variations can be observed between the higher-ranked samples (rank 1 and 5) and the lower ranked-sample (rank 100) in the nose and lip regions of image (a), the eye region of image (b), and the lip and chin region of image(c). This demonstrated the effectiveness of ranking in selecting higher-quality samples.}
    \label{fig:supp_ranking}
\end{figure}

We show additional expression variations from the final reconstruction of our method in~\cref{fig:supp_HardOcc} and \cref{fig:aflw2000_all}. For identity reconstruction,~\cref{fig:supp_ranking} presents more results from the ranking of samples evaluated on the NoW validation dataset. While most of the reconstructions appear visually similar, the variations are subtle (see the forehead patterns of rows (b) and (c)). In row (c), the chin area of least ranked sample shows high error compared to rank-1 and rank-5 samples. Since these subtle differences are hard to differentiate visually, ranking provides a way automatically select high-quality samples without manual intervention.

\begin{figure*}
    \centering
    \includegraphics[width=0.90\linewidth]{figures/results/supp_exp_2} 
    \caption{\textbf{Comparison of expression sampling on hard occlusions.} We compare against EMOCA~\cite{danecek2022emoca} (pink), three samplings from Diverse3D~\cite{dey2022generating} (blue) and 16 samples from our method (green).}
    \label{fig:supp_HardOcc}
\end{figure*}
\end{document}